\documentclass[a4paper,conference]{IEEEtran}
\IEEEoverridecommandlockouts

\usepackage{cite}
\ifCLASSINFOpdf
   \usepackage[pdftex]{graphicx}
\else
   \usepackage[dvips]{graphicx}
\fi
\usepackage{graphicx}

\usepackage{url}

\usepackage{booktabs}
\usepackage{color}
\usepackage{subfigure}
\usepackage[T1]{fontenc}

\newtheorem{lemma}{Lemma}

\def\DEL#1{\iffalse{#1}\fi}
\def\ADD#1{#1}

\begin{document}
%
\title{Agreement or Disagreement in Noise-tolerant Mutual Learning?}

\author{
	\IEEEauthorblockN{
		Jiarun Liu, Daguang Jiang, Yukun Yang,  
		Ruirui Li\thanks{\IEEEauthorrefmark{1} Corresponding author.}\IEEEauthorrefmark{1}}
	\IEEEauthorblockA{
		Information Science and Technical Institute\\
		Beijing University of Chemical Technology\\
		Beijing, China\\
		ilydouble@gmail.com}
}



%


\maketitle

\begin{abstract}
Deep learning has made many remarkable achievements in many fields but suffers from noisy labels in datasets. The state-of-the-art learning with noisy label method Co-teaching and Co-teaching+ confronts the noisy label by mutual-information between dual-network. However, the dual network always tends to convergent which would weaken the dual-network mechanism to resist the noisy labels. In this paper, we proposed a noise-tolerant framework named MLC in an end-to-end manner. It adjusts the dual-network with divergent regularization to ensure the effectiveness of the mechanism. In addition, we correct the label distribution according to the agreement between dual-networks. The proposed method can utilize the noisy data to improve the accuracy, generalization, and robustness of the network. We test the proposed method on the simulate noisy dataset MNIST, CIFAR-10, and the real-world noisy dataset Clothing1M. The experimental result shows that our method outperforms the previous state-of-the-art method. Besides, our method is network-free thus it is applicable to many tasks. \ADD{Our code can be found at \url{https://github.com/JiarunLiu/MLC.}}
\end{abstract}


%
\IEEEpeerreviewmaketitle

\section{Introduction}
Deep Neural Networks (DNN) have achieved remarkable success on many tasks. Most applied algorithms rely on accurate labels to supervise model learning. However, collecting extensive data and accurate labels is a daunting and laborious task. Many researchers turn to costless methods to acquire labels, such as online queries or crowdsourcing, which inevitably introduces label noise in the dataset and leads to a significant degeneration in the generalization of DNN models. 

Research on learning with noisy labels (LNL) has become a hot topic and strategies such as noise transition matrix\cite{patrini_making_2017}, redesigning loss functions\cite{ghosh_robust_2017,zhang_generalized_2018,miyato_virtual_2019}, or re-weighting samples\cite{ren_learning_2018,liu_classification_2016} et.al. are often combined with to obtain robust models. Recently, methods derived from co-training have achieved state-of-the-art results on the LNL tasks due to their ability to reduce error accumulation. Specifically, Co-teaching\cite{han_co-teaching_2018} maintains two networks simultaneously. Each network views its small-loss instances as useful knowledge and teaches such useful instances to its peer network for updating parameters. Researchers later found that mutual updates could cause the parameters of the two networks to converge, resulting in premature stopping of learning. To solve this problem, inspired by Decoupling\cite{malach_decoupling_2017}, Co-teaching+\cite{yu_how_2019} selects small-loss samples with inconsistent predictions between the two networks to mutually update the gradients. However, Co-teaching+ probably fails without sufficient qualified instances, especially when there is a large proportion of noisy labels. We have observed that Co-teaching+ performs worse than Co-teaching in some LNL tasks, such as medical image classification\cite{liu_co-correcting_2021}. On the contrary, JoCoR\cite{wei_combating_2020} believes that there is no need to ensure the inconsistency of the two networks. It uses a regularization term in the loss function to enhance the consistency of the predictions against the label noise. Therefore, there arises a question to be answered: Is “Disagreement” necessary for training two networks to deal with noisy labels?
\begin{figure}
	\centering
	\includegraphics[width=0.45\textwidth]{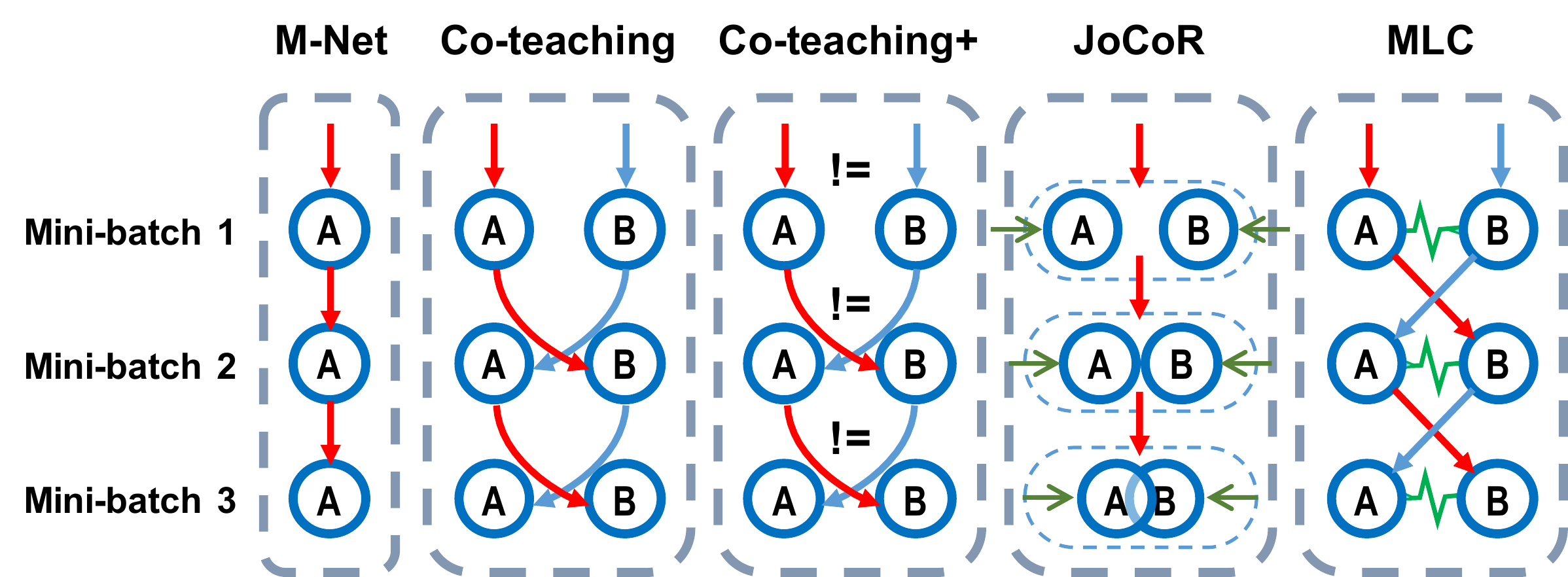}
	\caption{Comparison of different error flow. First Panel: MentorNet\cite{jiang_mentornet_2018} (M-Net) maintains only one network; Second Panel: Co-teaching\cite{han_co-teaching_2018} maintains two networks A and B. In each mini-batch, each network samples its small-loss instances and teaches these instances to its peer network; Third Panel: Co-teaching+\cite{yu_how_2019} adopts the idea of update by disagreement when selects useful instances; Fourth Panel: JoCoR\cite{wei_combating_2020} narrows the distance between A and B through a co-regularization on instances; Fifth Panel: MLC follows Co-teaching’s strategies but adds a co-regularization directly on network parameters to maintain a certain degree divergence.}
	\label{fig:main}
\end{figure}

To address this question, we conducted a series of exploration experiments and have observed that:
\begin{enumerate}
	\item Mutual learning tends to maximize prediction consistency.
	\item A certain disagreement is required for mutually training two networks to handle noisy labels.
\end{enumerate}

Based on the observations, there is an urgent need for us to come up with a mechanism capable of self-adjusting to maintain the optimal divergence between two networks. Regularization is a technique widely used in co-training, e.g., co-regularization in JoCoR. Unlike JoCoR, which updates gradients respectively and increases agreement by co-regularization, we design a new co-regularization from the perspective of maintaining disagreement between the two networks and propose a noise-tolerant learning framework \DEL{called}\ADD{named} Mutual Label Correction (MLC). To this end, we introduce a new way to measure the distance between the two networks. We also present a label probability distribution model based on mutual learning, which can be used in correcting the noisy labels along with the model training. As far as we can observe, MLC not only prevents premature agreement (like Co-teaching \ADD{in Fig. \ref{fig:acc_dis}}) but also avoids performing poorly on extremely noisy data (like Co-teaching+ \ADD{in Table \ref{tab:acc_mnist} and \ref{tab:acc_cifar10}}). Compared with JoCoR, MLC inherits the merits of mutual gradient updates which can ensure consistency and meanwhile better prevent error accumulation. Below, we summarize the key contributions of this work:
\begin{itemize}
	\item We discovered that mutual learning-based LNL methods need to maintain a certain degree of inconsistency while pursuing the consistency of predictions.
	\item A new co-regularization term for constraining the network divergence is designed to solve the problem of network convergence.
	\item A novel noise-tolerant learning framework with a new co-regularization is introduced, which adopts a label probability model and corrects the noisy labels during training.
\end{itemize}

\section{Related Work}

\subsection{Learning with noisy labels}
Many approaches have been proposed to the learning-with-noisy-labels (LNL) problem based on supervised learning\cite{vahdat_toward_2017} including designing new robust loss functions\cite{ghosh_robust_2017,zhang_generalized_2018}, estimating transition matrix\cite{patrini_making_2017} and etc. DNNs have been proved to have a certain degree of generalization and have been leveraged to promote classification robustness\cite{zhang_understanding_2021}. Some methods \cite{hendrycks_using_2018} require a small clean sample set for training, which is not suitable for many practical problems. \cite{arpit_closer_2017} has shown that DNNs tend to learn simple patterns first before fitting noises, resulting in a general rule that samples with lower loss are more likely to be clean samples. Based on this, \cite{arazo_unsupervised_2019} fits a two-component Beta Mixture Model (BMM) to distinguish the distribution of clean and noisy samples, which tends to produce undesirable flat distributions under asymmetric noises. MentorNet\cite{jiang_mentornet_2018} leverages curriculum learning to screen correct labels for student network training. Inspired by Co-training, Co-teaching\cite{han_co-teaching_2018} and Co-teaching+\cite{yu_how_2019} train two neural networks simultaneously and learn each other's small loss samples to update together. JoCoR\cite{wei_combating_2020} does not uses cross-updating. Instead, it calculates a joint loss with Co-Regularization and uses small-loss instances in updating gradients to explicitly reduce the diversity of two networks. DivideMix\cite{li_dividemix_2019} combines two networks in implicit and explicit two ways. It models samples loss distribution with a Gaussian Mixture Model (GMM) to divide the dataset into a labeled set (mostly clean) and an unlabeled set (mostly noisy), then uses an improved MixMatch\cite{berthelot_mixmatch_2019} method to perform semi-supervised training. \ADD{Unlike previous work, MoPro\cite{li_mopro_2020} combined self-supervised learning and supervised learning together to release the unscalability in webly-supervised representation learning.}

\subsection{Co-training}
As an extension of self-training, co-training\cite{blum_combining_1998} lets multiple individual models iteratively learn from each other\cite{zhou_semi-supervised_2010,wang_theoretical_2017}. It requires the “sufficiency and independence” assumptions hold. That is to say two classifiers should keep diverged to achieve the better ensemble effects. Recently, Deep Co-training\cite{qiao_deep_2018} maintains disagreement through a view difference constraint. For example, to prevent consensus, TriNet\cite{chen_tri-net_2018} designs different head structures and samples different sub-datasets for learners. \cite{li_comatch_2021} applies a graph-based regularization and also integrates supervised contrastive learning. However, in the tasks of learning with noisy labels, without the distinct views of data, co-training has to promote diversity in some other ways.

\section{Preliminary} \label{sec:preliminary}
Inspired by co-training in semi-supervised learning, Co-teaching, Co-teaching+, and JoCoR all adopt the scheme of training two networks simultaneously. But they follow different error flow strategies to combat label noise, just as Fig. \ref{fig:main} illustrated. Both Co-teaching and Co-teaching+ adopt cross updates of mutual learning. 

In order to further understand the performance of mutual learning on noisy labels, we conducted exploratory experiments \DEL{on three noisy datasets, namely MNIST with 20\% pair-flip pollution, CIFAR-10 with 20\% pair-flip pollution, and MNIST with 80\% random pollution. }\ADD{on two noisy datasets, namely MNIST and CIFAR-10 with 20\% pair-flip pollution.} We use KL divergence to measure the degree of inconsistency between the two networks in predicted probability distributions, and define the divergence as follows:
\begin{equation}
	Divergence = KL(p_1,p_2) + KL(p_2,p_1)
\end{equation}
where $p_1$ is the Softmax output of the first network, $p_2$ is that of the second network, and $KL()$ is the operation of KL divergence. 
\begin{figure}
        \centering
        \subfigure[MNIST Co-teaching]{
        \includegraphics[width=4cm]{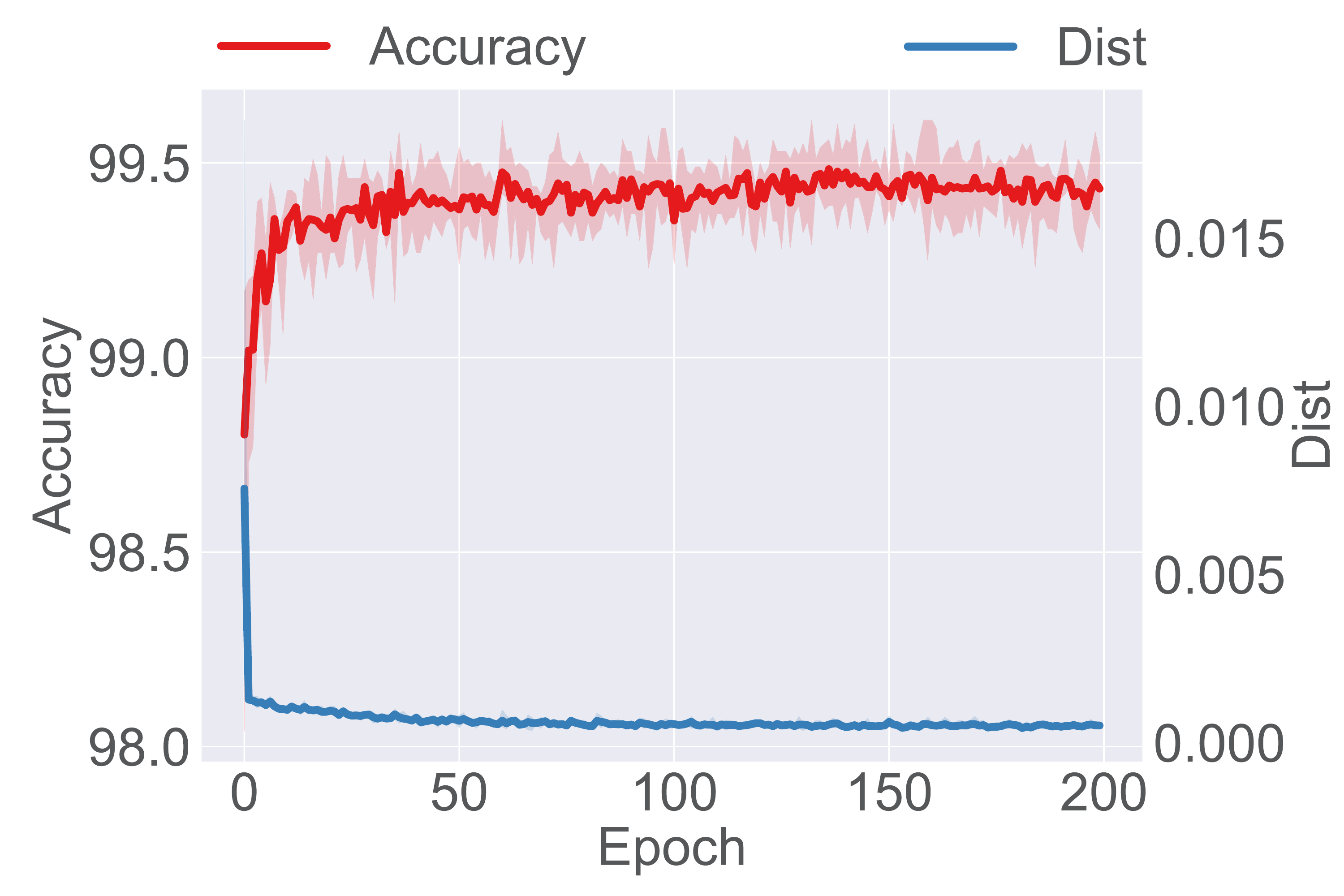} \label{fig:acc_dis_a}
        }  
        \subfigure[CIFAR-10 Co-teaching]{
        \includegraphics[width=4cm]{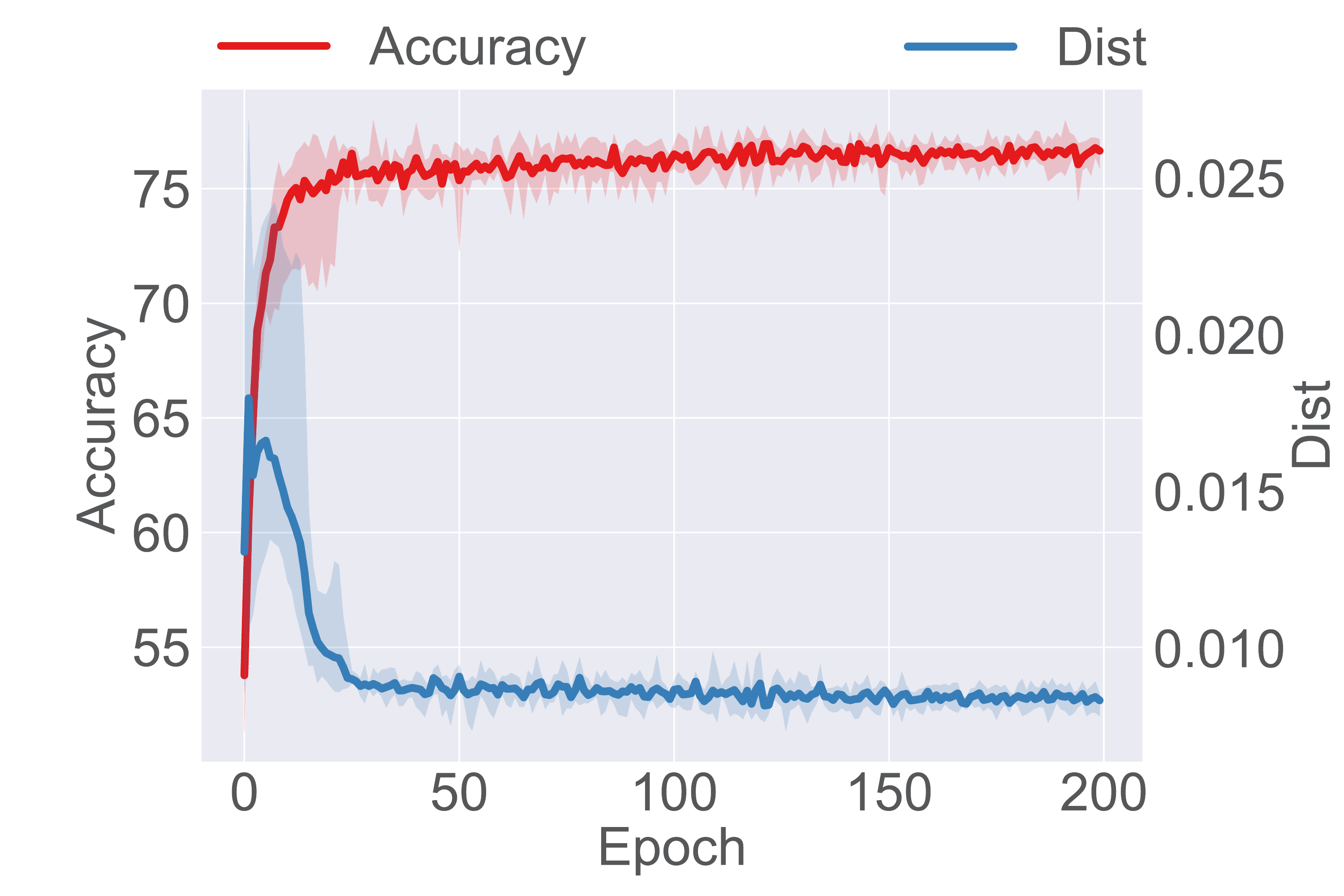} \label{fig:acc_dis_b} 
        }
        \subfigure[MNIST Co-teaching w/o]{
        \includegraphics[width=4cm]{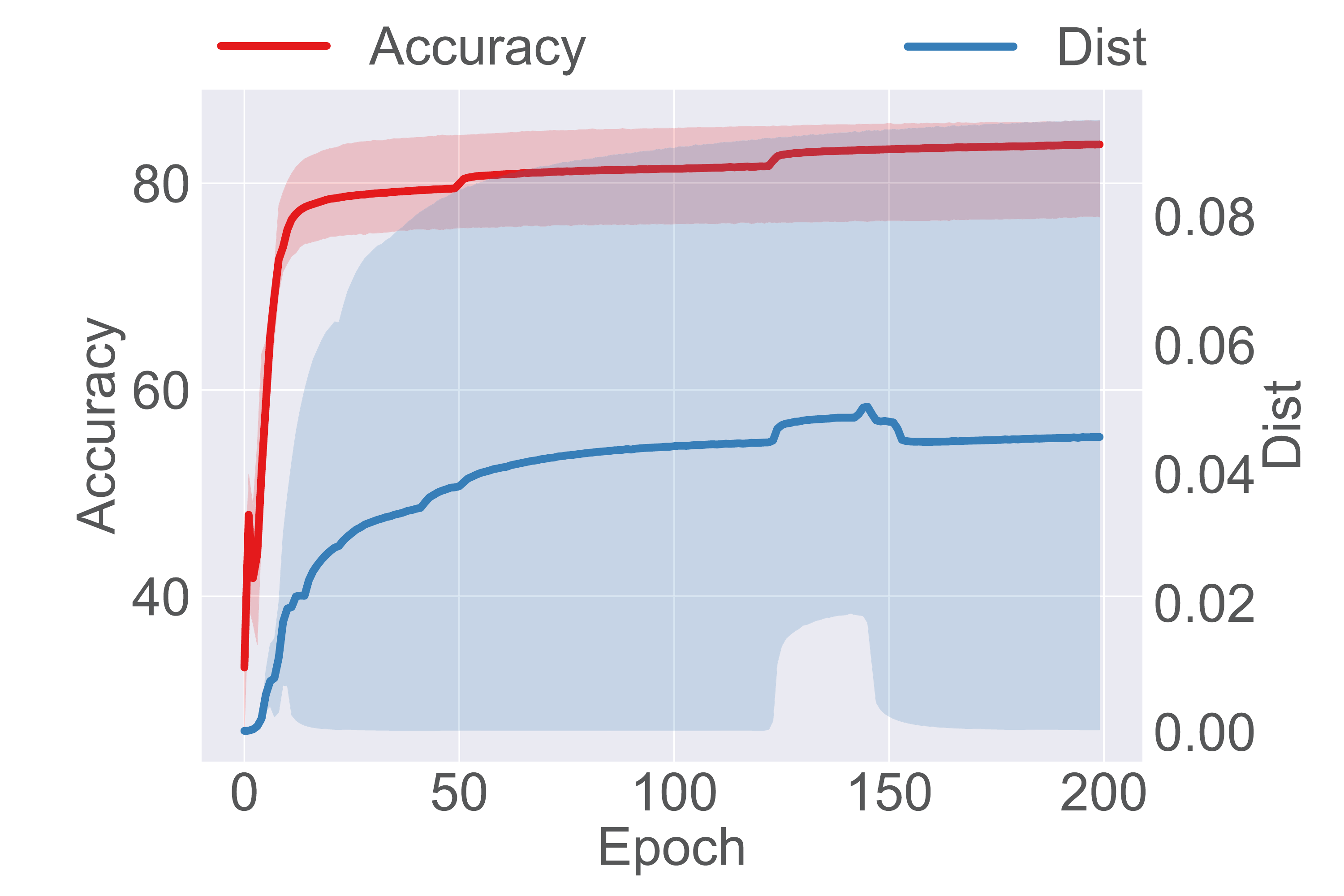} \label{fig:acc_dis_c}
        }
        \subfigure[CIFAR-10 Co-teaching w/o]{
        \includegraphics[width=4cm]{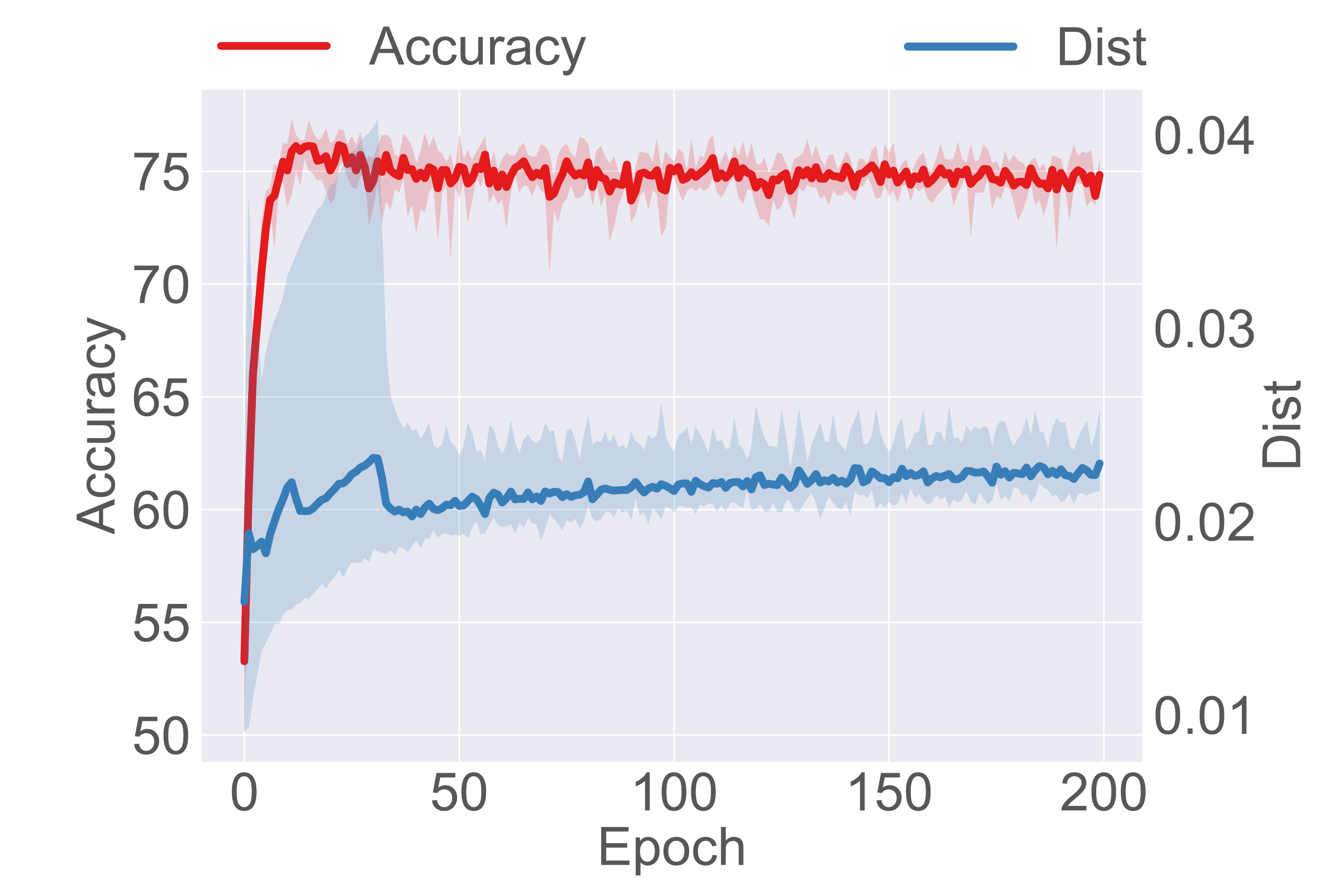} \label{fig:acc_dis_d}
        }  
        \caption{Accuracy and Divergence of Co-teaching with or without mutual updates on MNIST and CIFAR-10 datasets with $20\%$ pair-flip noise.}
        \label{fig:acc_dis}
\end{figure}

Fig.\ref{fig:acc_dis} reports the \ADD{exploratory experiments result of the} total variation in divergence and accuracy on the test sets. The red line represents the overall accuracy of the current model, and the blue line represents the divergence of the predictions between the two networks. It could be found that Co-teaching that uses mutual updates (Fig. \ref{fig:acc_dis_a}, \ref{fig:acc_dis_b}) has better robustness and generalization than training two networks independently (Fig. \ref{fig:acc_dis_c}, \ref{fig:acc_dis_d}). For Co-teaching, as the training process progresses, the overall accuracy increases while the value of divergence decreases. This suggests that mutual updates will make the two networks increasingly consistent. In essence, noisy labels exacerbate the complexity of the optimization space, easily causing gradient-based optimizers to fit noise and get stuck in local optima. Mutual information updates from the peer network will reduce the likelihood of getting stuck in local minima, but its ultimate goal is to form consistent predictions in the global optimum. So we get the observation as follows.
\begin{lemma}
	Mutual learning tends to maximize prediction consistency.
\end{lemma}

\begin{figure}
	\centering
	\subfigure[\ADD{CIFAR-10 sn-0.2}]{
        \includegraphics[width=4cm]{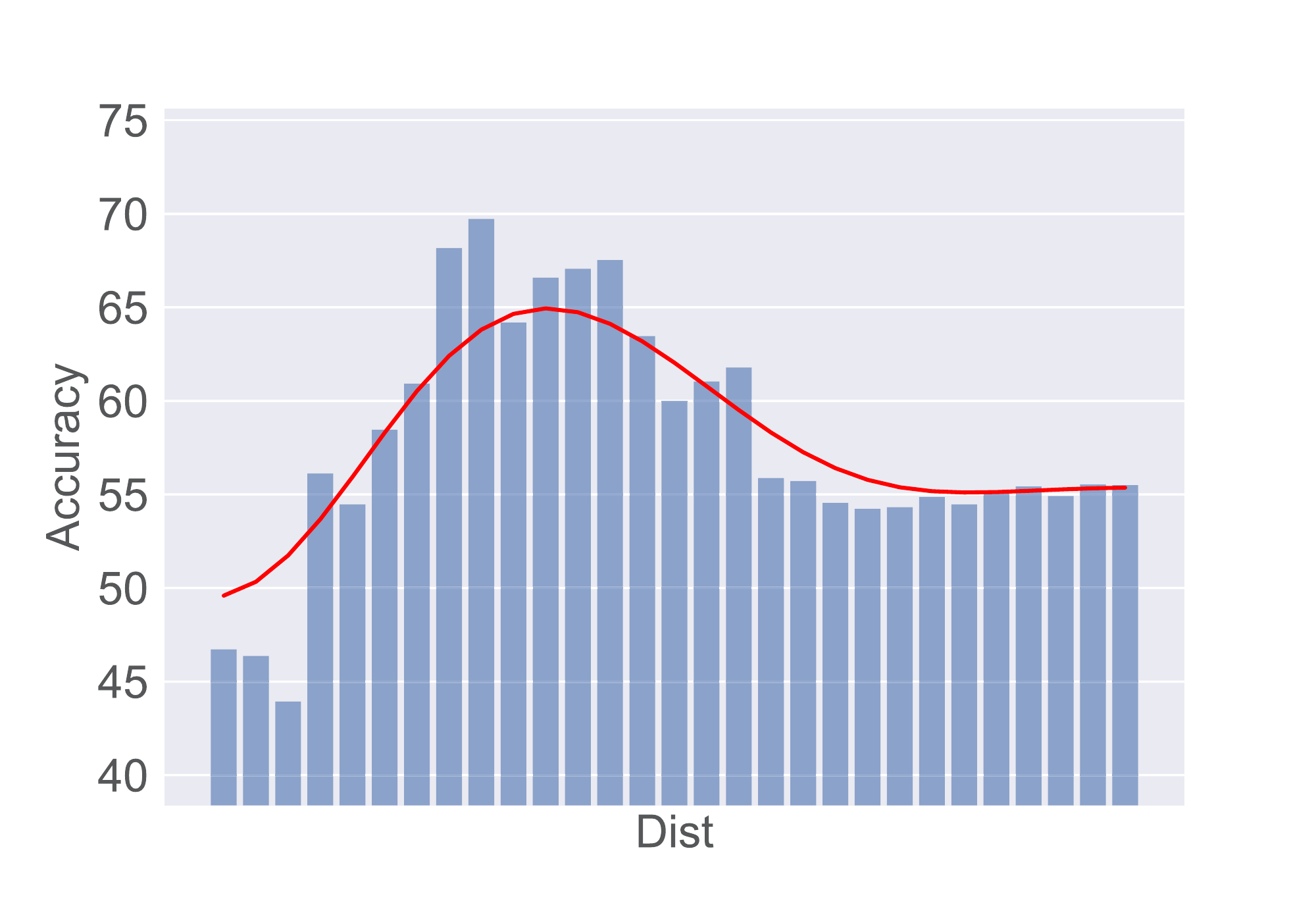}
    }
	\subfigure[\ADD{CIFAR-10 sn-0.4}]{
        \includegraphics[width=4cm]{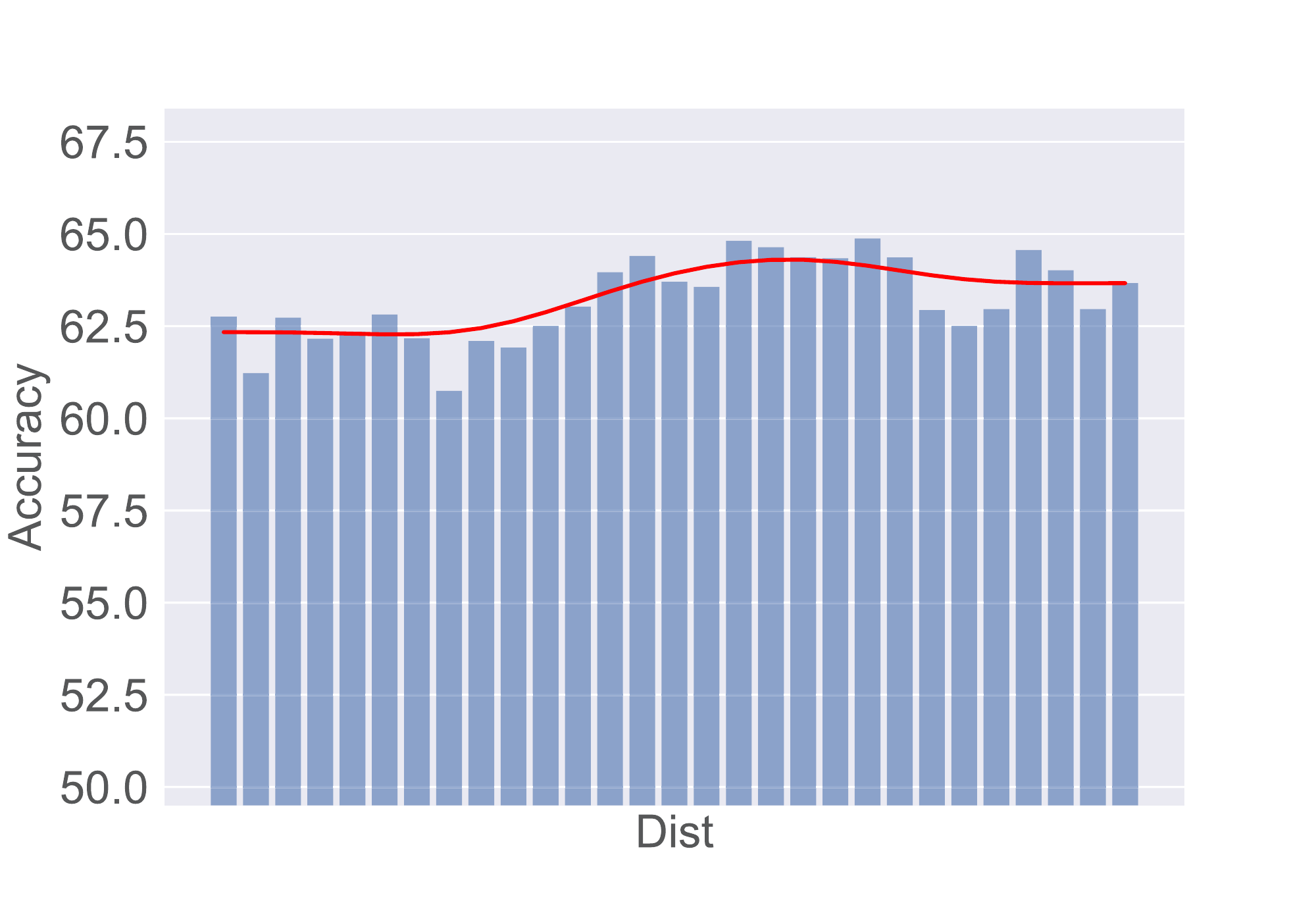}
    }
	\subfigure[\ADD{CIFAR-10 sn-0.8}]{
        \includegraphics[width=4cm]{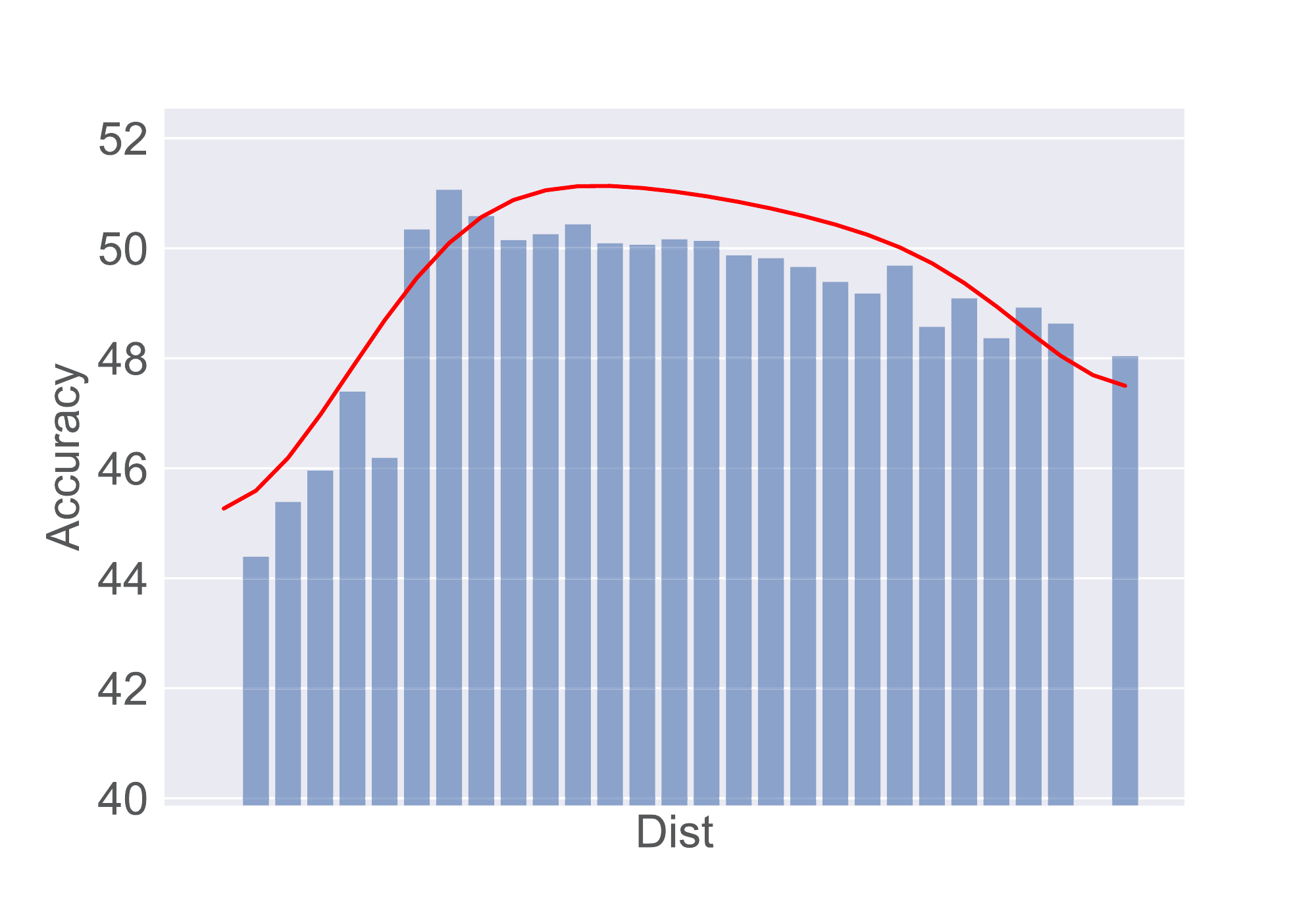}
    }
	\subfigure[\ADD{MNIST sn-0.2}]{
        \includegraphics[width=4cm]{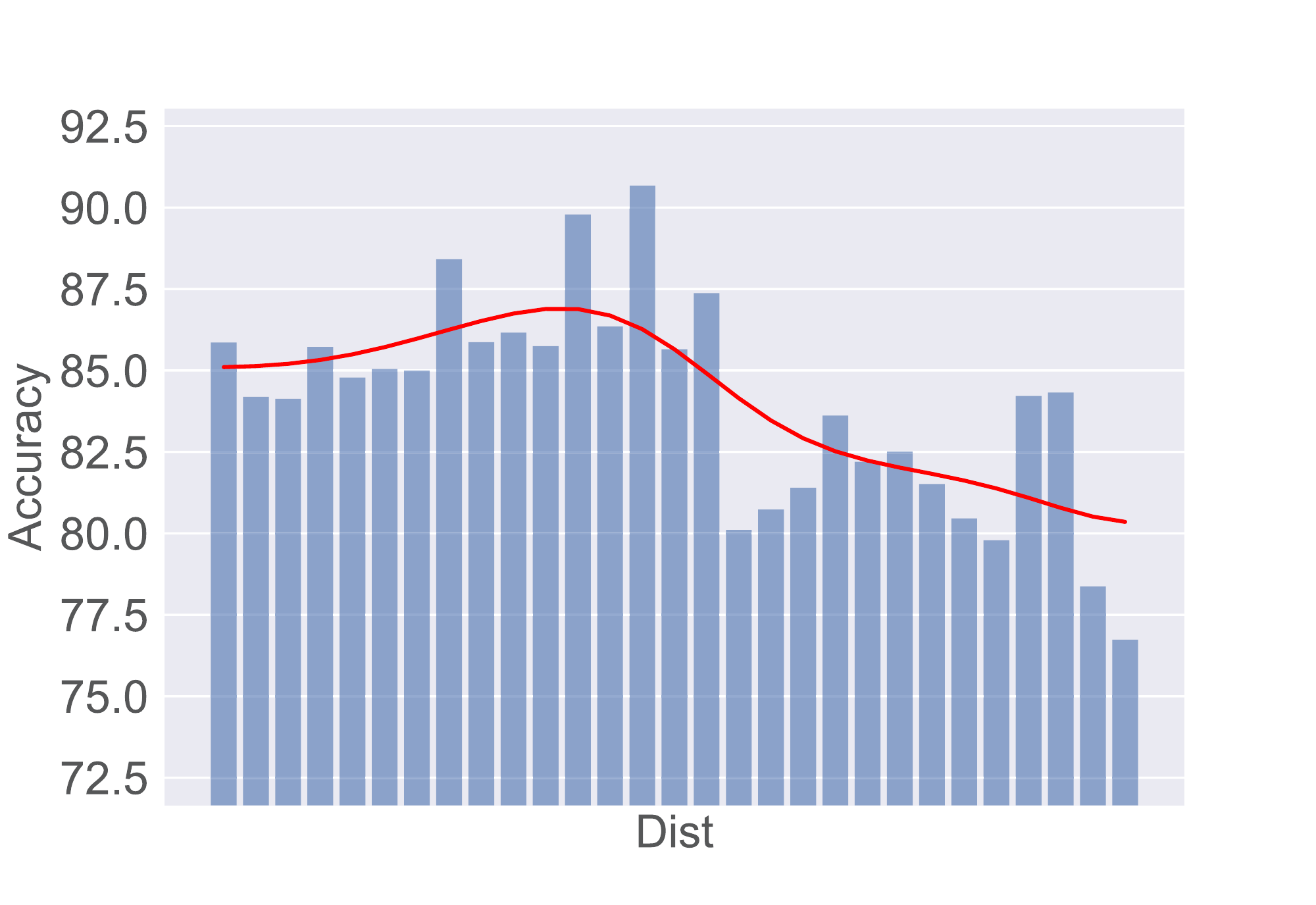}
    }
	\subfigure[\ADD{MNIST sn-0.4}]{
        \includegraphics[width=4cm]{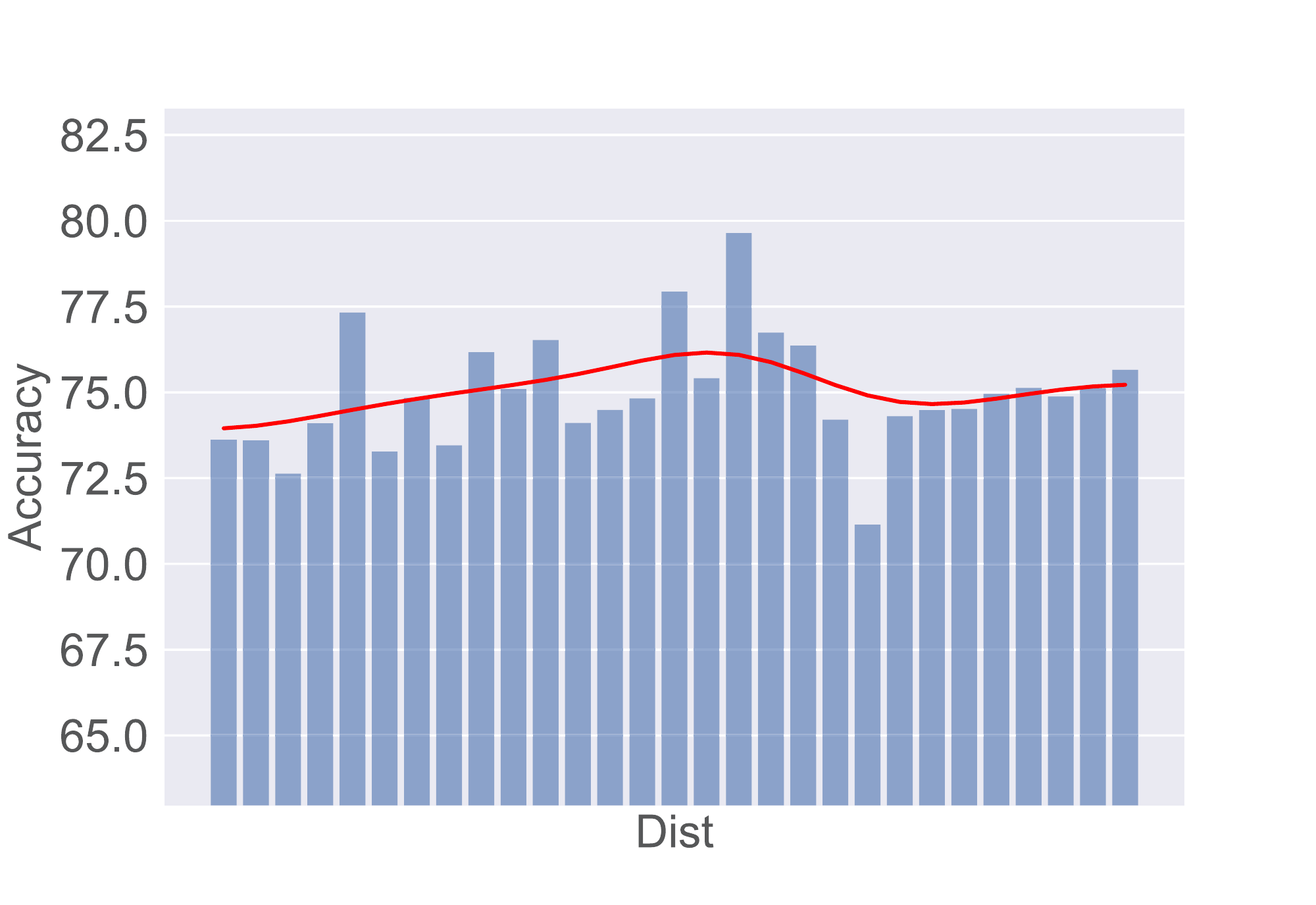}
    }
	\subfigure[\ADD{MNIST sn-0.8}]{
        \includegraphics[width=4cm]{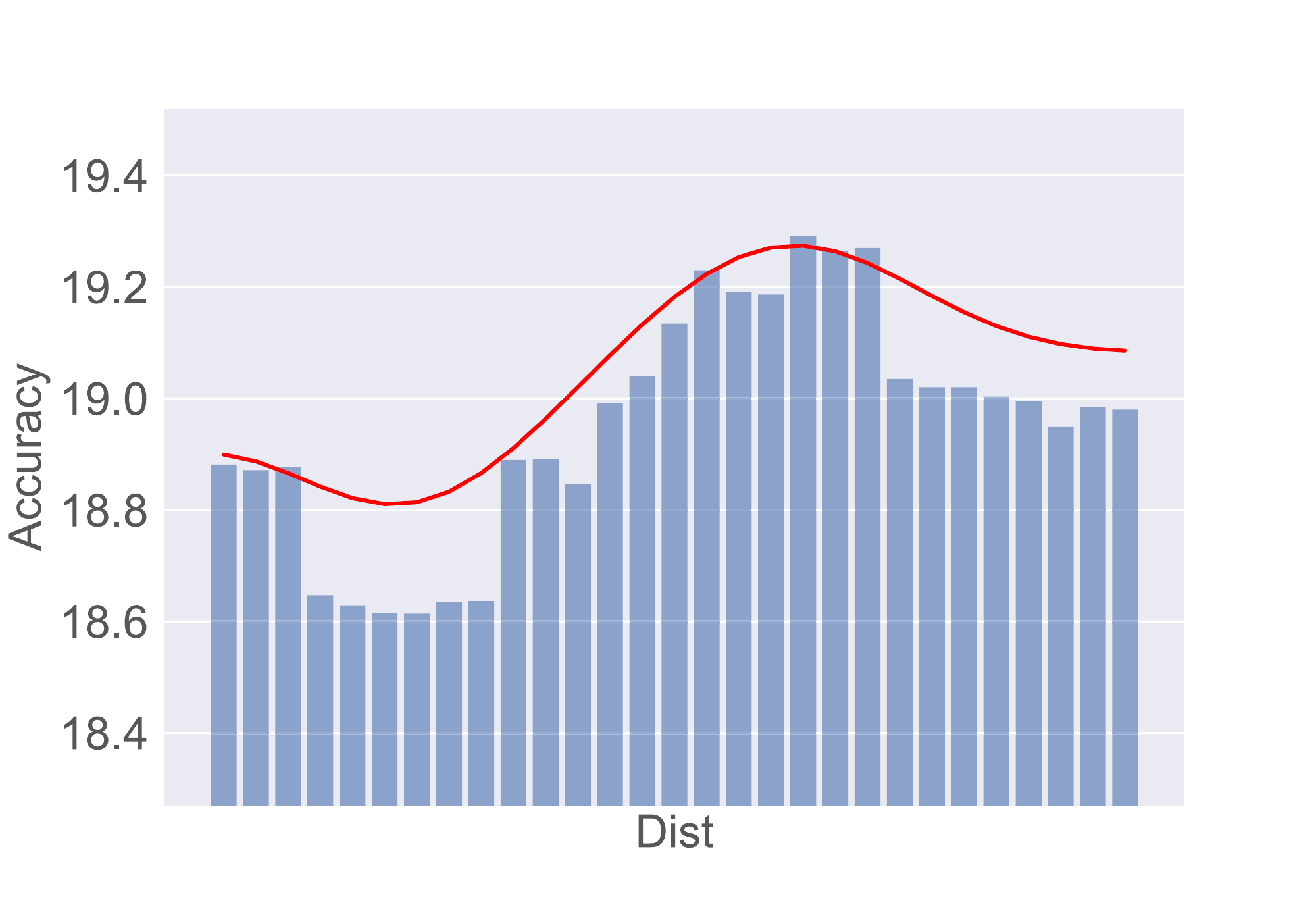}
    }
	\caption{\ADD{The variation of average accuracy under different network divergences. }}
	\label{fig:variation}
\end{figure}

However, during mutual training, two networks may converge prematurely due to the unbalanced sample distribution, which will cause Co-teaching to degenerate into self-paced learning\cite{yu_how_2019} Co-teaching+ tries to tackle this problem by adopting an “update by disagreement” strategy, which results in slower and more volatile optimization. Fig. \ref{fig:variation} shows the experimental results of Co-teaching on \DEL{MNIST with 80\% random pollution}\ADD{MNIST and CIFAR-10 with different random label corruption}. It shows the variation of average accuracy under different network divergences. There is a clear trend that with the divergences increase they first go up and then come down \ADD{within different dataset and noise corruption}.  It indicates that a certain degree of variance helps to better judge label accuracy on noisy datasets, thereby improving generalization. Then we get the second observation as follows.
\begin{lemma}
	A certain disagreement is required for mutually training two networks to handle noisy labels.
\end{lemma}

Therefore, there is a need for us to find a better way to update the gradients and design an effective mechanism to maintain the divergence of different networks\ADD{, thus to maintain the robustness via mutually training}.

\section{Proposed Method}

\subsection{Co-Regularization}
Many co-training-based methods add regularization to make predictions less divergent, ensuring that networks will eventually learn a consistent global optimum. Unfortunately, it will introduce bias locally. Especially in the case of noisy annotations, the situation becomes complicated and difficult to quantify. 
To avoid this kind of bias, we design an adaptive co-regularization from the perspective of quantifying the distance of network parameters. This is because network has the ability to remember. Its parameters are more stable and less irrelevant to label noise. 
Given two collaborative networks $f\left(\theta_1\right)$ and $f\left(\theta_2\right)$ parameterized by $\theta_1,\theta_2$, we use Euclidean distance to measure the distance of network parameters as follows:
\begin{equation}
	dist=\Vert \theta_1,\theta_2 \Vert_2
\end{equation}

And the co-regularization term $l_d$ can be formulate as:
\begin{equation}
	l_d={dist}^\mu
\end{equation}
where $\mu$ is a negative hyper-parameter. A lower $\mu$ forces models to separate more. As previously mentioned in Section \ref{sec:preliminary}, MLC adopts mutual updates which will naturally let models from different networks get closer during training. Therefore, in this paper, we hope to prevent the consensus between the two networks by setting this regularization. Specifically, in our experiment we set $\mu=-1$. It is worth noting that the learner will degenerate to Co-teaching when $\mu=-\infty$, and it will degenerate to co-teaching+ when $\mu=0$.

\subsection{Mutual Label Correction Framework}
\begin{figure}
	\centering
	\includegraphics[width=0.45\textwidth]{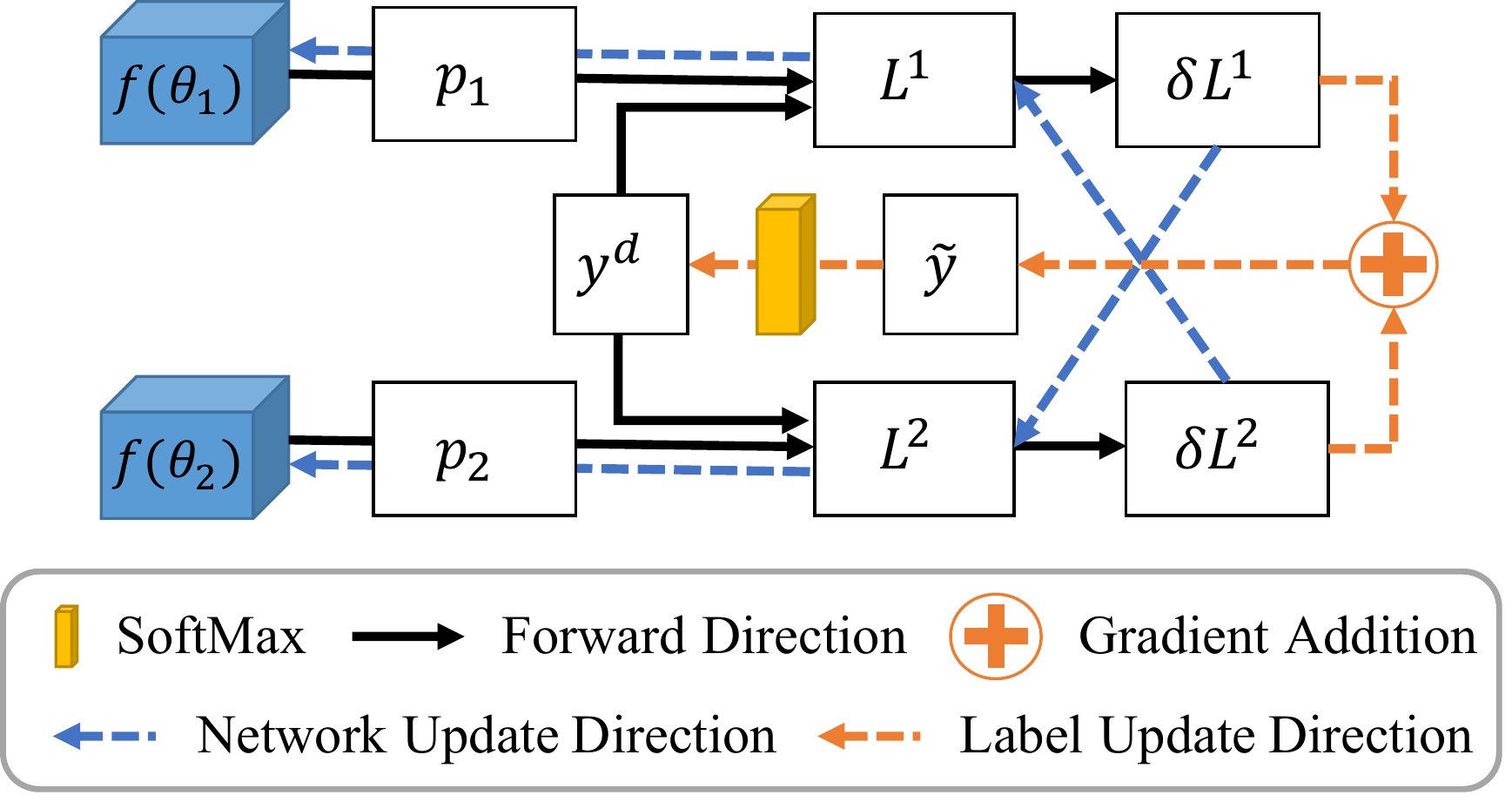}
	\caption{Mutual Label Correction (MLC) architecture.}
	\label{fig:mlc_arch}
\end{figure}

Based on mutual learning and co-regularization, we proposed a mutual label correction (MLC) framework for noise-tolerant classification. The key architect is illustrated in Fig. \ref{fig:mlc_arch}. It contains two kinds of learning at the same time, namely network update (dotted blue lines) and label update (dotted orange lines). 

We adopt the label probability distribution model $y^d$ that was originally proposed by PENCIL\cite{yi_probabilistic_2019} and adapt it for our collaborative training. To this end, we define an auxiliary variable $\widetilde{y}$ to assist the label updating which is initialized by multiplying one-hot encoding of the noisy labels $\hat{y}$ and a constant hyper-parameter $K$:
\begin{equation}
	\widetilde{y}=K\hat{y}
\end{equation}

Then the label probability distribution $y^d$ can be obtained through a softmax function:
\begin{equation}
	y^d=softmax(\widetilde{y})
\end{equation}

By inputting $p_1, p_2$ and $y^d$ into the loss function which is defined in the following section, we can obtain the losses $l_1$ and $l_2$ respectively. Then we compute the gradient of them with respect to the network parameters and label probability distribution through back-propagation. The No.1 network is updated by the gradient of $\frac{\delta L^2}{\theta_2}$ and the No.2 network is updated by the gradient of $\frac{\delta L^1}{\theta_1}$. 

In PENCIL, since it has only a single label distribution, the way to update the label distribution is as follows:
\begin{equation}
	\widetilde{y}\gets\widetilde{y}-\lambda\frac{\delta L}{y^d}
\end{equation}

where $\lambda$ is the step size of the label update. To maximize label consistency, we update the labels using the sum of the two network gradients:
\begin{equation}
	\widetilde{y}\gets\widetilde{y}-\lambda(\frac{\delta L^1}{y^d}+\frac{\delta L^2}{y^d})	
\end{equation}

\subsection{Loss Function}
The loss function of MLC contains multiple loss terms, which are defined as follows:
\begin{equation}
	L=l_c\left(p,y^d\right)+\alpha l_o\left(\hat{y},y^d\right)+\beta l_e\left(p\right)+\xi l_d\left(\theta_1,\theta_2\right)
\end{equation}
In which $l_c$ is flipped Kullback-Leibler (KL) divergence of prediction and label distribution. $l_o$ is the KL divergence between original noisy labels $\hat{y}$ and updated label distribution $y^d$.  $l_e$ is the Shannon entropy of network prediction to avoid over-smooth prediction. $l_d$ is co-regularization. $\alpha,\ \beta,\ \xi$ are their weighted hyper-parameters. Detailed formulation of $l_c, l_o$ and $l_e$ can be found at \cite{yi_probabilistic_2019}.

\section{Experiment}

\subsection{Experiment Setup}

\subsubsection{Datasets}
We compare MLC with state-of-the-art methods on two classical datasets with artificial label noise and a real-world noisy dataset, Clothing1M.
\begin{figure}
        \centering
        \subfigure[symmetric]{
        \includegraphics[width=4cm]{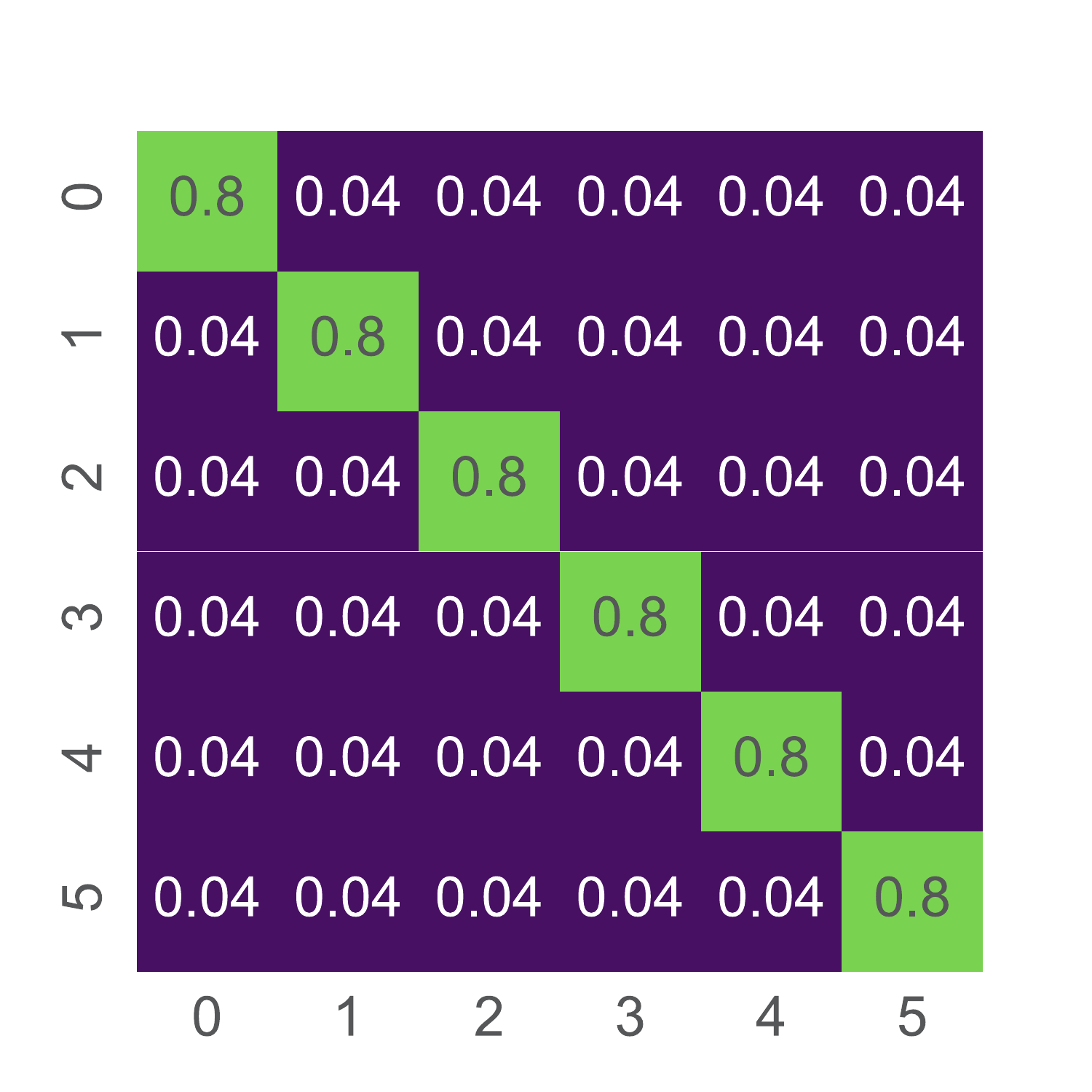}
        }  
        \subfigure[pairflip]{
        \includegraphics[width=4cm]{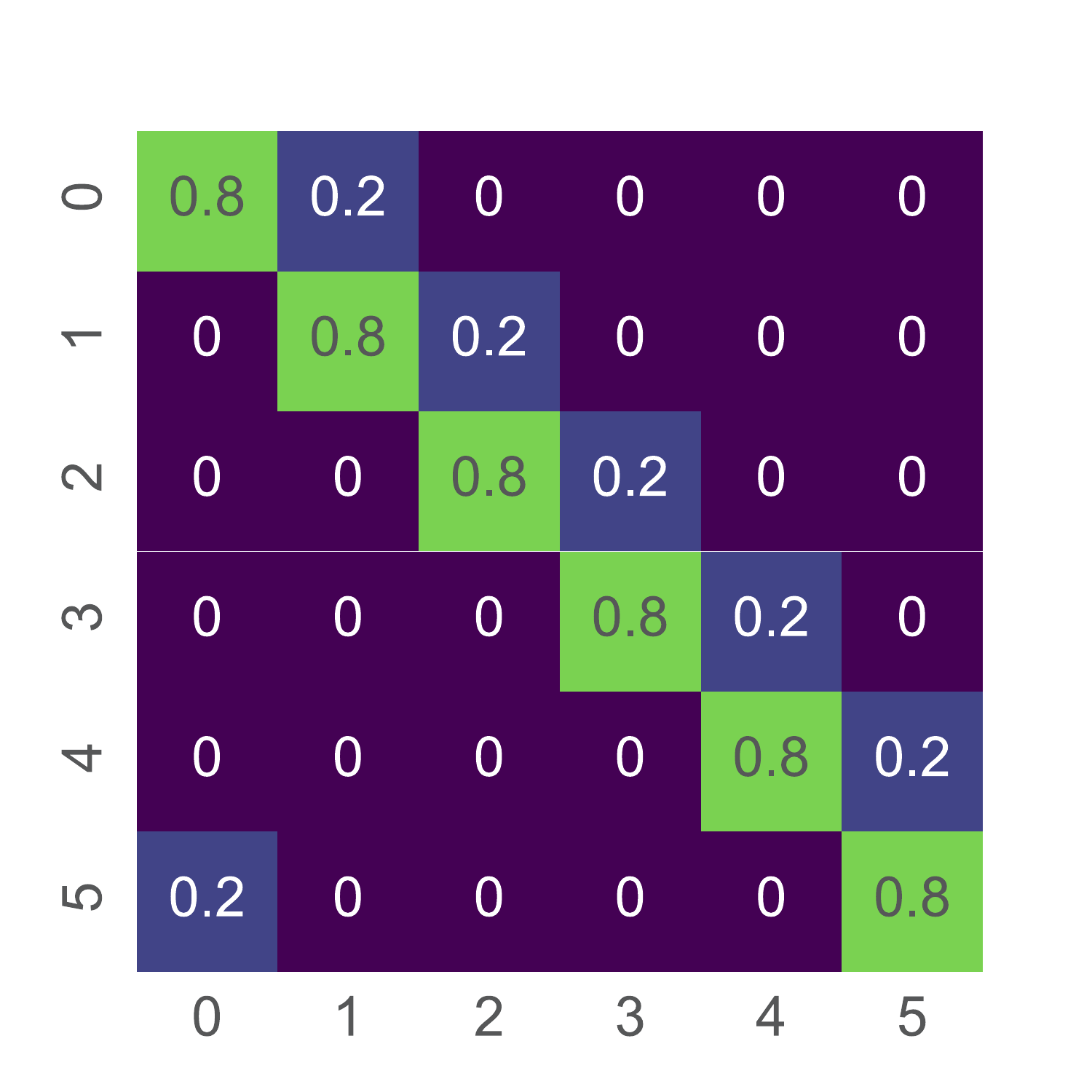}
        }
        \caption{Transition matrices of different noise types (using 6 classes as an example with $20\%$ noise)}
        \label{fig:trans_mat}
\end{figure}
\begin{table}[h]
  \centering
  \caption{Choice of label update step on MNIST and CIFAR-10.}
  \begin{tabular}{cccccc}
    \toprule
        & sn-0.2 & sn-0.4 & sn-0.8 & pair-0.2 & pair-0.45  \\
    \midrule
    MNIST    & 1000   & 3000   & 3000   & 2000     & 2500       \\
    CIFAR-10 & 3000   & 4000   & 4000   & 1500     & 4000       \\
    \bottomrule
  \end{tabular}
  \label{tab:lambda}
\end{table}

\textbf{MNIST} contains 60k handwriting characters of 0-9 with size 28px*28px. We split the dataset into a training set of 50k images and a testing set of 10k images. All images are randomly augmented by image perspective transformation and color jittering.

\textbf{CIFAR-10} contains 60k 32px * 32px images of 10 categories with 50k training data and 10k testing data. We apply random horizontal flips and random cropping to all training data.

According to the definition provided by \cite{han_co-teaching_2018} (Fig. \ref{fig:trans_mat}), we use symmetric noise (sn) and pairflip noise (pair) to randomly corrupt labels on MNIST and CIFAR-10. We test with symmetric noise of progressively increased noise ratios on these two datasets, including 0.2, 0.4, and 0.8. For the more challenging pairflip noise, we test with it in the ratios of 0.2 and 0.45.

\textbf{Clothing1M} contains 1 million training data with real-world label noise and 10526 clean testing data in 14 classes of different clothing types. Following JoCoR, we scale all images into 256px * 256px and apply 224px *224px centered crop. After that, we applied the normalization for all images.

Our experiments were carried out on the software platform of Ubuntu 18.04 LTS and PyTorch 1.4.0, using the hardware of Intel i7-8700k CPU, NVIDIA RTX2080Ti GPU, and 32GB RAM.

\begin{table*}[ht]
  \centering
  \caption{Accuracy on MNIST}
  \setlength{\tabcolsep}{4mm}
  \begin{tabular}{ccccccccccc}
    \toprule
                 & \multicolumn{2}{c}{\textbf{sn-0.2}}                               & \multicolumn{2}{c}{\textbf{sn-0.4}}                               & \multicolumn{2}{c}{\textbf{sn-0.8}}                               & \multicolumn{2}{c}{\textbf{pair-0.2}}                             & \multicolumn{2}{c}{\textbf{pair-0.45}}                             \\
                 & \textit{best}                   & \textit{last}                   & \textit{best}                   & \textit{last}                   & \textit{best}                   & \textit{last}                   & \textit{best}                   & \textit{last}                   & \textit{best}                   & \textit{last}                    \\
    \midrule
    \textit{Standard}     & 97.29 & 91.04              & 96.45 & 83.86              & 86.69 & 35.14              & 96.28 & 85.1                 & 73.13 & 55.72                  \\
    \textit{Co-teaching}  & 97.56 & 97.45              & 96.5  & 96.24              & 80.43 & 76.87              & 96.7  & 96.22                & 90.29 & 88.4                   \\
    \textit{Co-teaching+} & 96.82 & 96.64              & 95.84 & 95.7               & 75.13 & 74.49              & 96.73 & 96.44                & 82.21 & 81.61                  \\
    \textit{PENCIL}       & 97.31 & 93.94              & 96.44 & 85.31              & 86.58 & 33.44              & 96.07 & 85.61                & 74.52 & 55.08                  \\
    \textit{JoCoR}        & 97.23 & 97.2               & 96.16 & 96.11              & 76.62 & 74.35              & 96.27 & 96.15                & 90.63 & 90.18                  \\
    \textit{\ADD{MoPro}}        & 97.21 & 96.75              & 96.46 & 96.22              & 83.65 & 82.84              & 97.11 & 96.83                & 66.97 & 64                     \\
    \midrule
    \textit{MLC}          & \textbf{\textcolor{red}{97.92}} & \textbf{\textcolor{red}{97.85}} & \textbf{\textcolor{red}{97.52}} & \textbf{\textcolor{red}{97.43}} & \textbf{\textcolor{red}{89.86}} & \textbf{\textcolor{red}{89.58}} & \textbf{\textcolor{red}{97.95}} & \textbf{\textcolor{red}{97.85}} & \textbf{\textcolor{red}{91.56}} & \textbf{\textcolor{red}{90.28}}  \\
    \bottomrule             
  \end{tabular}
  \label{tab:acc_mnist}
\end{table*}
\begin{table*}[h]
  \centering
  \caption{Accuracy on CIFAR-10}
  \setlength{\tabcolsep}{4mm}
  \begin{tabular}{ccccccccccc}
    \toprule
                          & \multicolumn{2}{c}{\textbf{sn-0.2}}                               & \multicolumn{2}{c}{\textbf{sn-0.4}}                               & \multicolumn{2}{c}{\textbf{sn-0.8}}                               & \multicolumn{2}{c}{\textbf{pair-0.2}}                             & \multicolumn{2}{c}{\textbf{pair-0.45}}                             \\
                          & \textit{best}                   & \textit{last}                   & \textit{best}                   & \textit{last}                   & \textit{best}                   & \textit{last}                   & \textit{best}                   & \textit{last}                   & \textit{best}                   & \textit{last}                    \\
   \midrule                       
    \textit{Standard}     & 73.86                           & 64.97                           & 67.58                           & 45.09                           & 35.48                           & 15.36                           & 72.84                           & 70.68                           & 50.8                            & 44.09                            \\
    \textit{Coteaching}   & 76.11                           & 74.08                           & 72.06                           & 69.37                           & 30.97                           & 16.92                           & 74.78                           & 73.61                           & 56.06                           & 52.78                            \\
    \textit{Co-teaching+} & 75.66                           & 69.84                           & 72.03                           & 51.07                           & 31.01                           & 13.65                           & 74.66                           & 67.61                           & 56.4                            & 41.32                            \\
    \textit{PENCIL}       & 75.95                           & 67.72                           & 71.14                           & 50.73                           & 35.72                           & 14.69                           & 75.34                           & 70.69                           & 51.77                           & 46.8                             \\
    \textit{JoCoR}        & 82.51                           & 82.22                           & 77.19                           & 76.94                           & 23.19                           & 22.19                           & 81.17                           & 80.68                           & 48.59                           & 45.08                            \\
    \textit{\ADD{MoPro}}        & 77.7                            & 76.07                           & 70.04                           & 59.68                           & 50.84                           & 44.23                           & 81.55                           & 80.67                           & 57.75                           & 50.63                            \\
    \midrule    
    \textit{MLC}          & \textbf{\textcolor{red}{84.59}} & \textbf{\textcolor{red}{84.46}} & \textbf{\textcolor{red}{80.66}} & \textbf{\textcolor{red}{80.45}} & \textbf{\textcolor{red}{41.88}} & \textbf{\textcolor{red}{41.48}} & \textbf{\textcolor{red}{84.71}} & \textbf{\textcolor{red}{84.56}} & \textbf{\textcolor{red}{65.23}} & \textbf{\textcolor{red}{64.16}}  \\
    \bottomrule
  \end{tabular}
  \label{tab:acc_cifar10}
\end{table*}
\begin{table*}[h]
  \centering
  \caption{Result on Clothing1M}
  \setlength{\tabcolsep}{6.7mm}
  \begin{tabular}{ccccccc}
    \toprule
    \textbf{Methods}  & \textit{Standard} & \textit{Co-teaching} & \textit{Co-teaching+} & \textit{PENCIL} & \textit{JoCoR} & \textit{MLC}                    \\
    \midrule   
    \textbf{Accuracy} & 64.68             & 68.51                & 58.79                 & 66.53           & 69.79          & \textcolor{red}{\textbf{69.8}}  \\
    \bottomrule
  \end{tabular}
  \label{tab:acc_cloth}
\end{table*}

\subsubsection{Baselines}
We compare MLC with the following state-of-the-art algorithms. All the codes for comparison were applied from the official implementation.

\textbf{Standard} This is the most basic training algorithm without any modification or tricks to combating noisy labels.

\textbf{Co-teaching} This is the first algorithm in combating noisy labels with dual-network. It combating noisy labels by mutual information exchange and small-loss sample selection.

\textbf{Co-teaching+} This is the algorithm based on the Co-teaching. It claims the "Update by disagreement" strategy by disagreement sample selection.

\textbf{PENCIL} This is the most representative algorithm of learning and correcting the label distribution simultaneously.

\textbf{JoCoR} This is the algorithm that claims the agreement updating by a joint regular term, which can reduce the distance between networks.

\ADD{\textbf{MoPro} This is the latest robust method that combine self-supervised learning and supervised learning together.}

\subsubsection{Network and Optimizer}
For the experiments on MNIST, we use a two-layer MLP as the backbone\cite{han_co-teaching_2018}. The batch size is set to 128 and uses Adam optimizer. For the 3-stage training, we train MLC in a total of 320 epochs, including 30 warmup epochs and 180 finetune epochs. The learning rate is 0.001 at the beginning 140 epochs and then decrease to 0.0001. The choice of label update parameter $\lambda$ is listed in Table \ref{tab:lambda}.

On CIFAR-10, we use a six-layer CNN backbone. We train MLC 320 epochs in total. The learning rate is 0.005 in 200 epochs and then decrease to 0.0001.

For the experiment on Clothing1M, we follow the setting of JoCoR, which has a 64 batch size of ResNet-18 backbone, optimized by Adam. The learning rate is 0.0001 and $\lambda$ is 4000.

\subsubsection{Measurement}
We use the classification accuracy to measure network performance:
\begin{equation}
Accuracy=\frac{number\ of\ correct\ prediction}{number\ of\ samples}
\end{equation}

Notably, we record the best test accuracy during training named as best and the average test accuracy of the last ten epochs named as last. In our experiments, it is not always that higher best accuracy is associated with high last accuracy. We perform five trials training on MNIST and CIFAR-10 to report their average results. Specifically, the comparison results of Clothing1M are referenced from \cite{wei_combating_2020}.

\subsection{Results}
Table \ref{tab:acc_mnist} and Table \ref{tab:acc_cifar10} show the classification accuracy of all comparison methods on MNIST and CIFAR-10. MLC won first place under various noise ratios/types \ADD{in most cases}. We found that noisy labels do degenerate the accuracy of DNN classifiers from experimental results. As the noise ratio increases, the accuracy of almost all the comparison methods decreases with varying degrees. For instance, the accuracy (last) of the standard method are reduced by more than 50\% when the noise ratio equals 0.8 on MNIST and CIFAR-10.

\begin{figure}
        \flushleft
        \subfigure[pairflip 0.2]{
        \includegraphics[width=4cm]{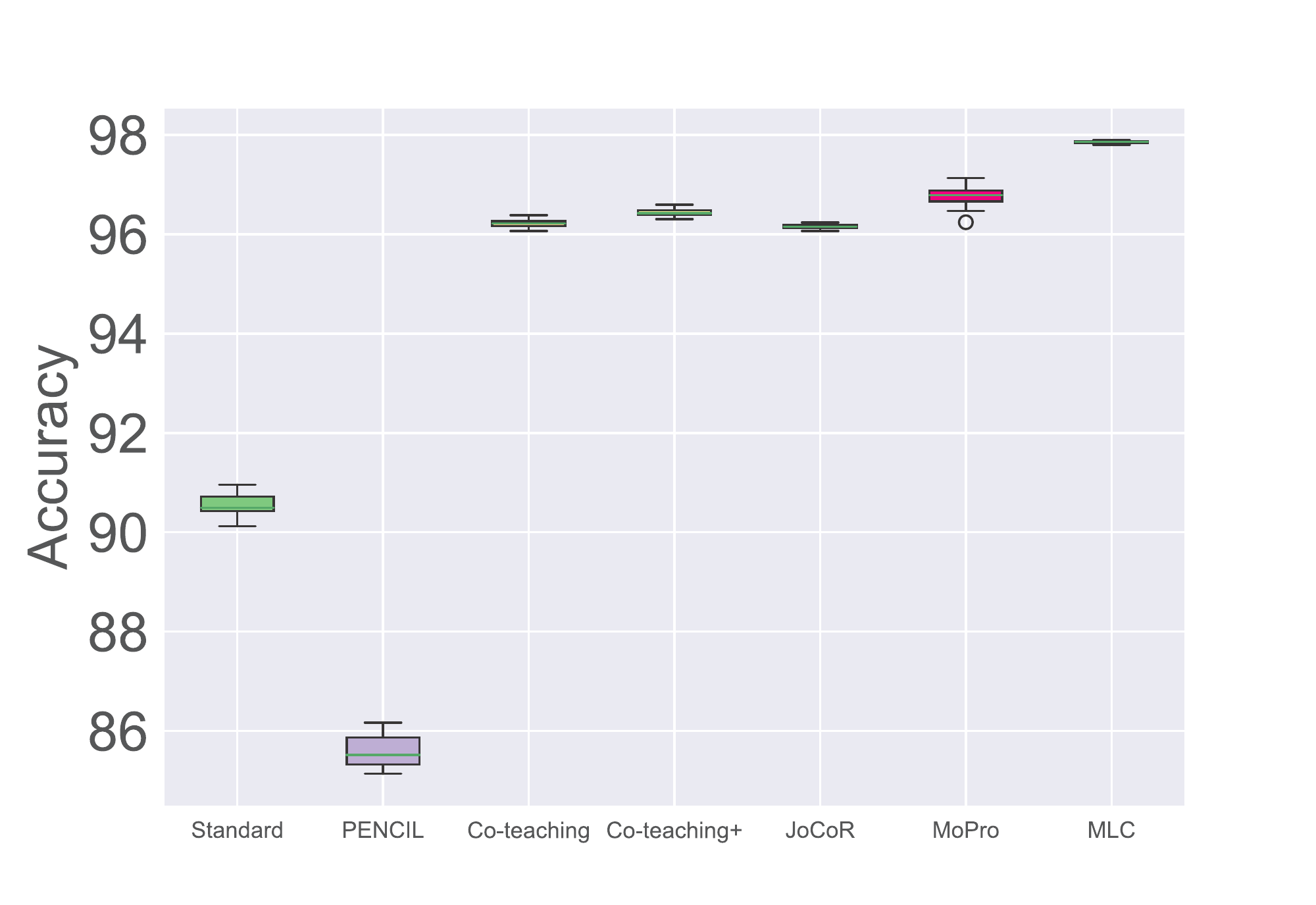} \label{fig:mnist_box_a}
        }  
        \subfigure[pairflip 0.45]{
        \includegraphics[width=4cm]{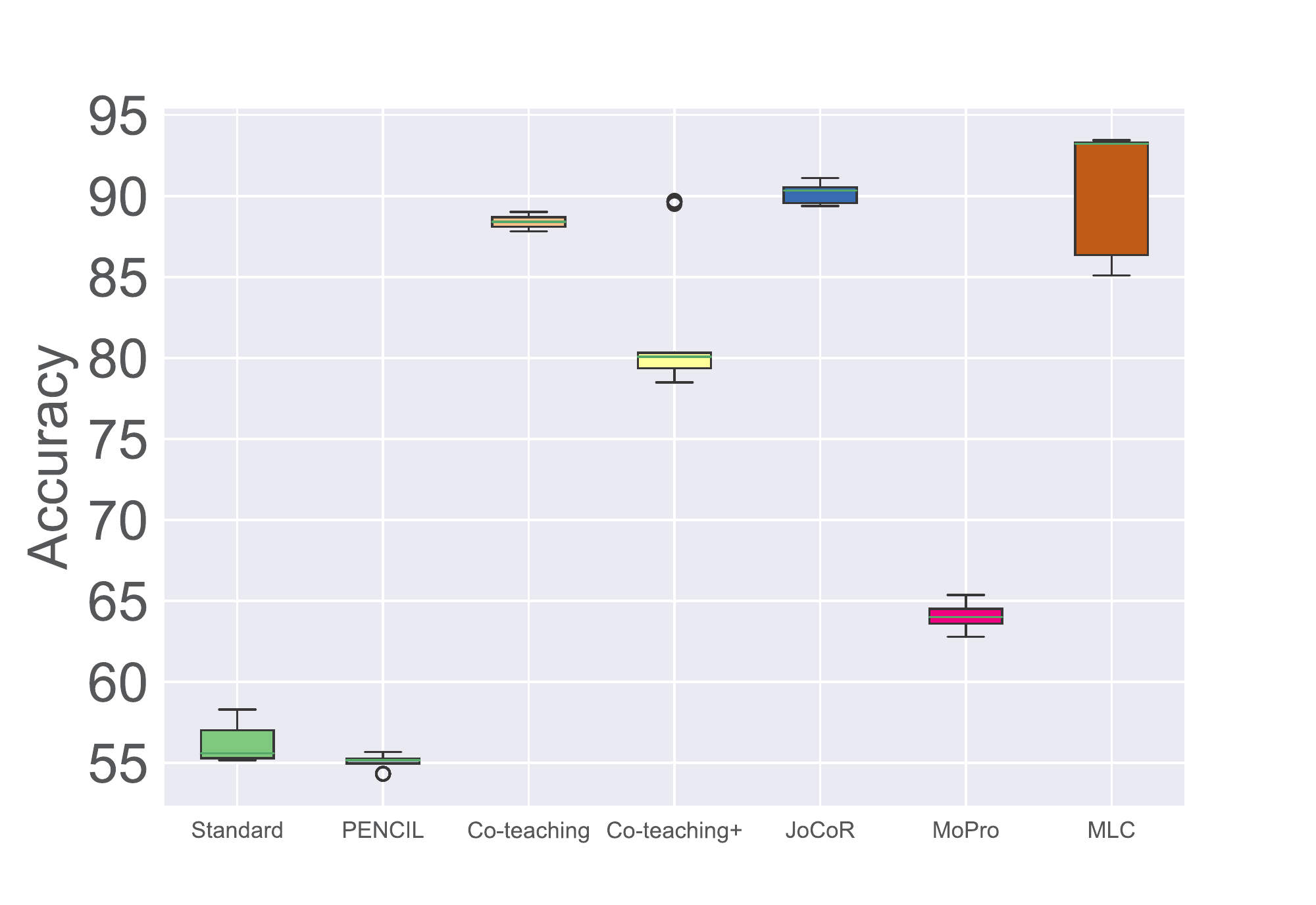} \label{fig:mnist_box_b}
        }
        \subfigure[symmetric 0.2]{
        \includegraphics[width=4cm]{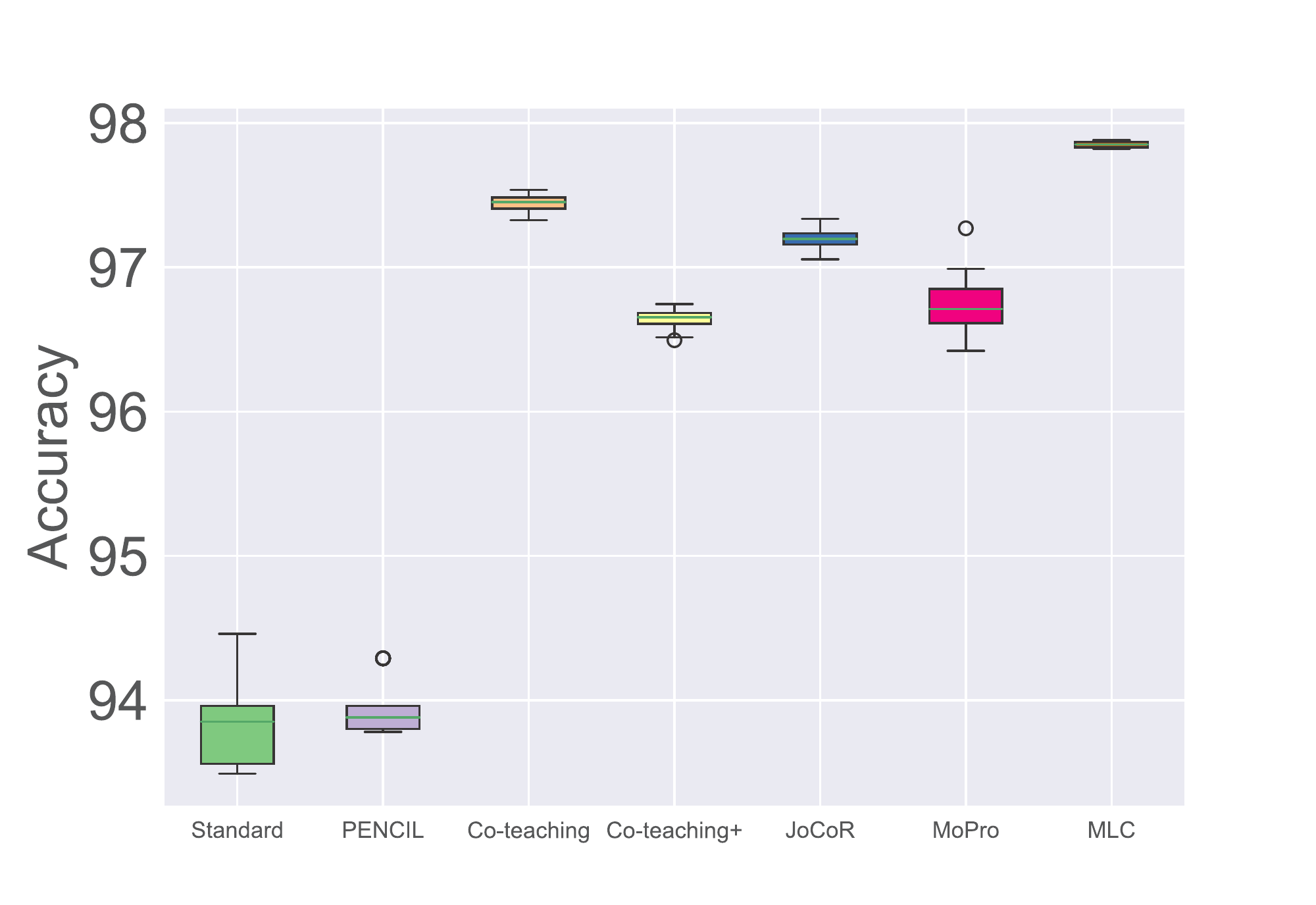} \label{fig:mnist_box_c}
        }
        \subfigure[symmetric 0.4]{
        \includegraphics[width=4cm]{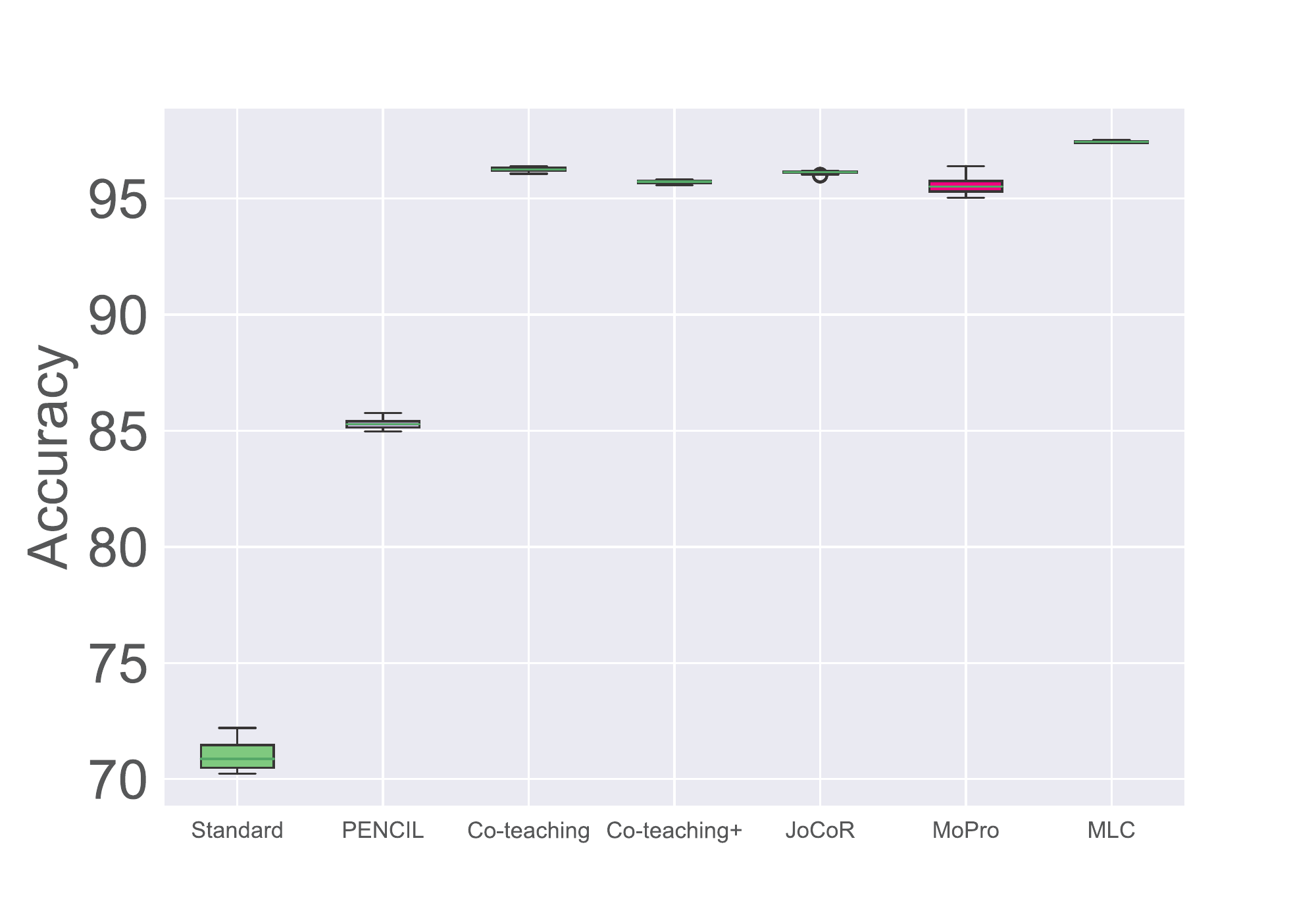} \label{fig:mnist_box_d}
        }
        \subfigure[symmetric 0.8]{
        \includegraphics[width=4cm]{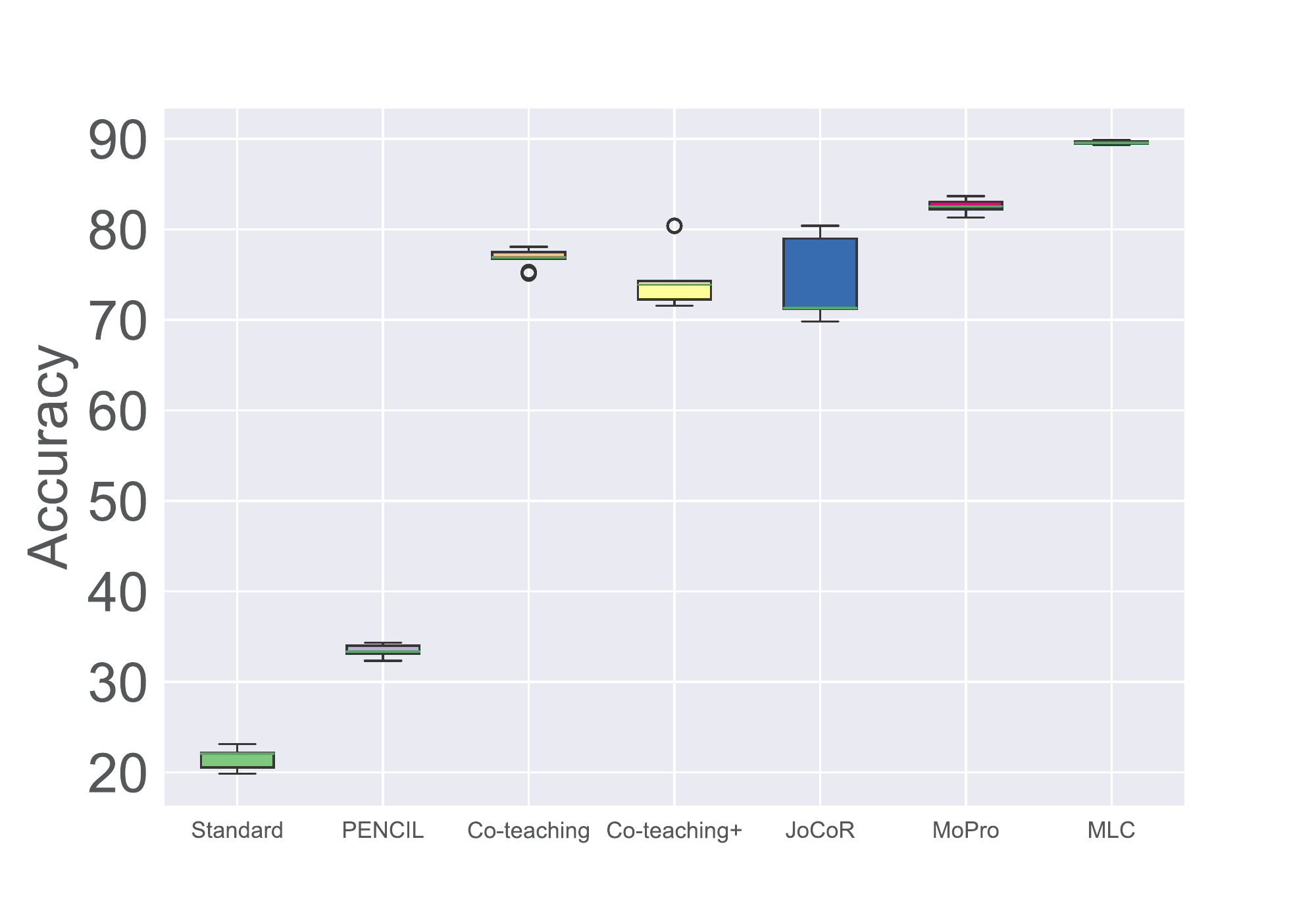} \label{fig:mnist_box_e}
        }
        \hspace{4mm}
        \subfigure{
        \includegraphics[width=3cm]{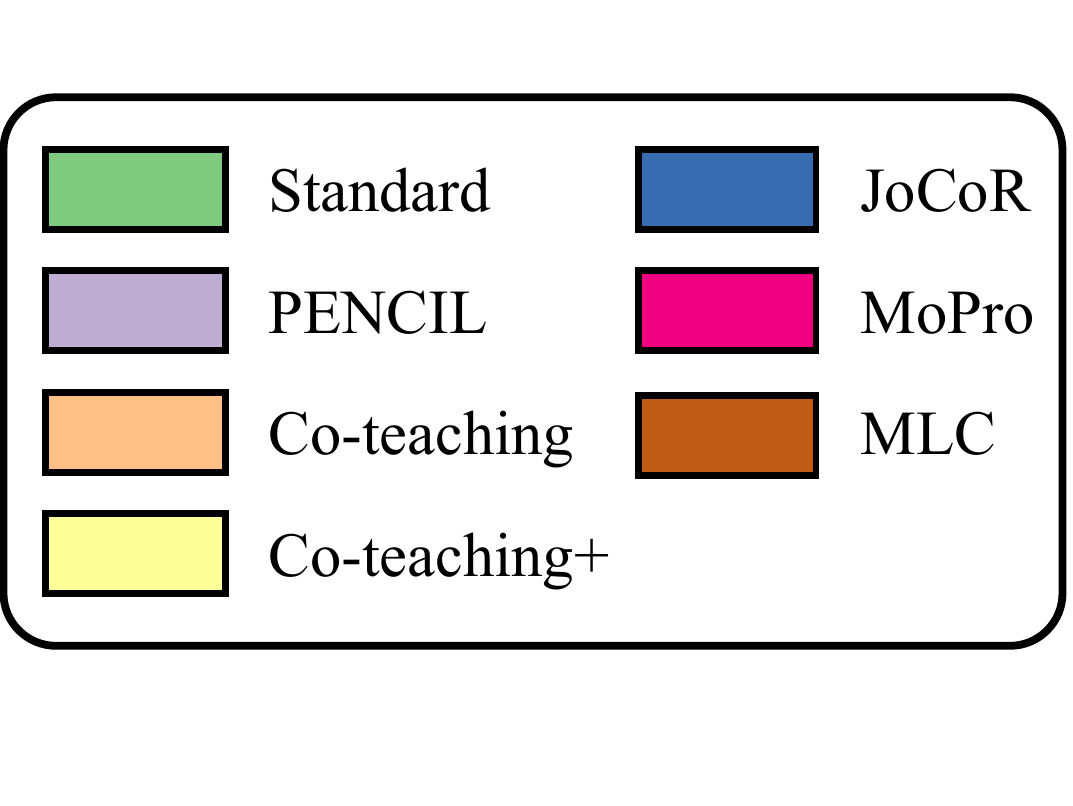}
        }
        \caption{Accuracy box-whisker plot of MNIST based on last ten epochs of five trials.}
        \label{fig:mnist_box}
\end{figure}

\begin{figure}
        \flushleft
        \subfigure[pairflip 0.2]{
        \includegraphics[width=4cm]{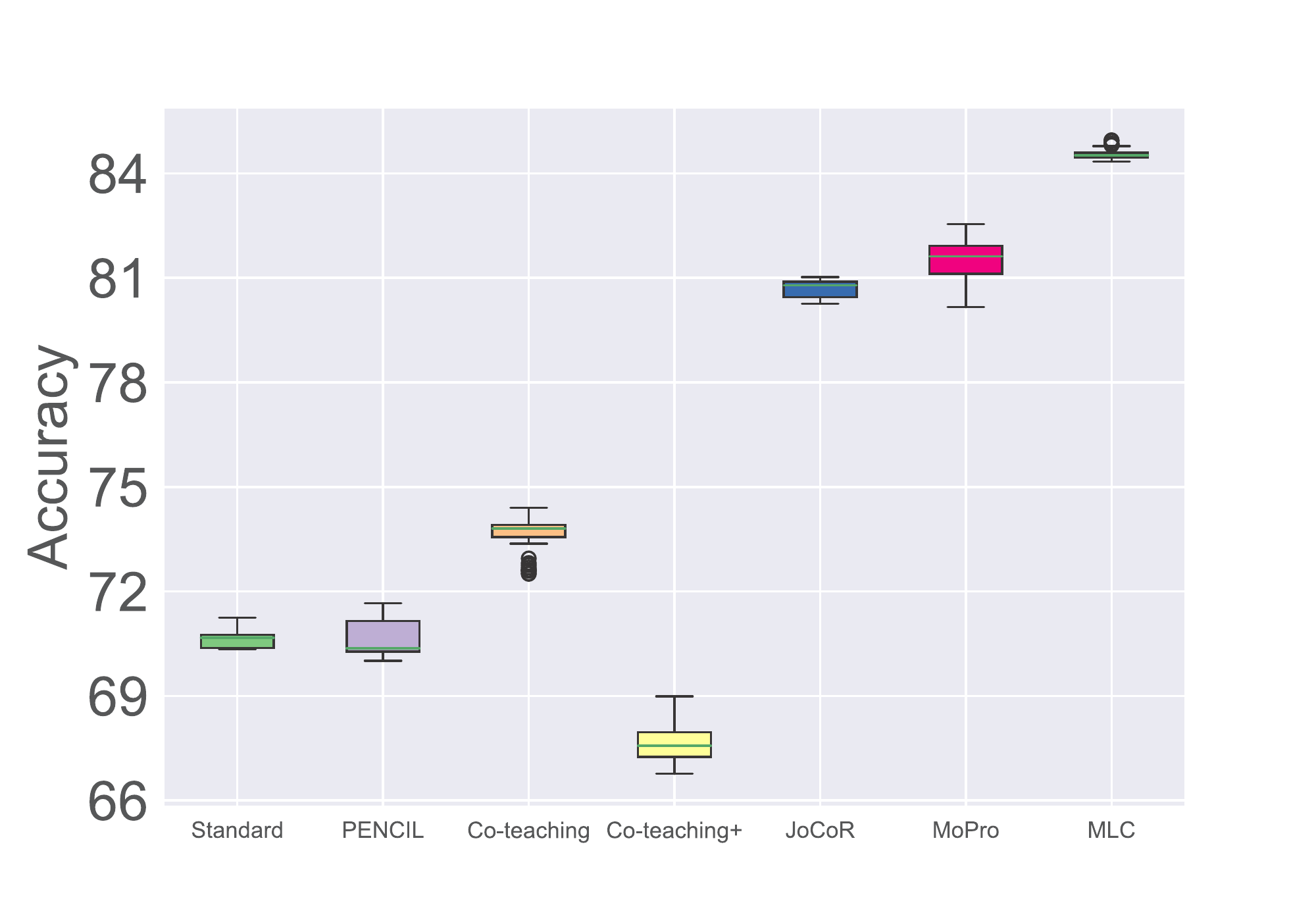} \label{fig:cifar10_box_a}
        }  
        \subfigure[pairflip 0.45]{
        \includegraphics[width=4cm]{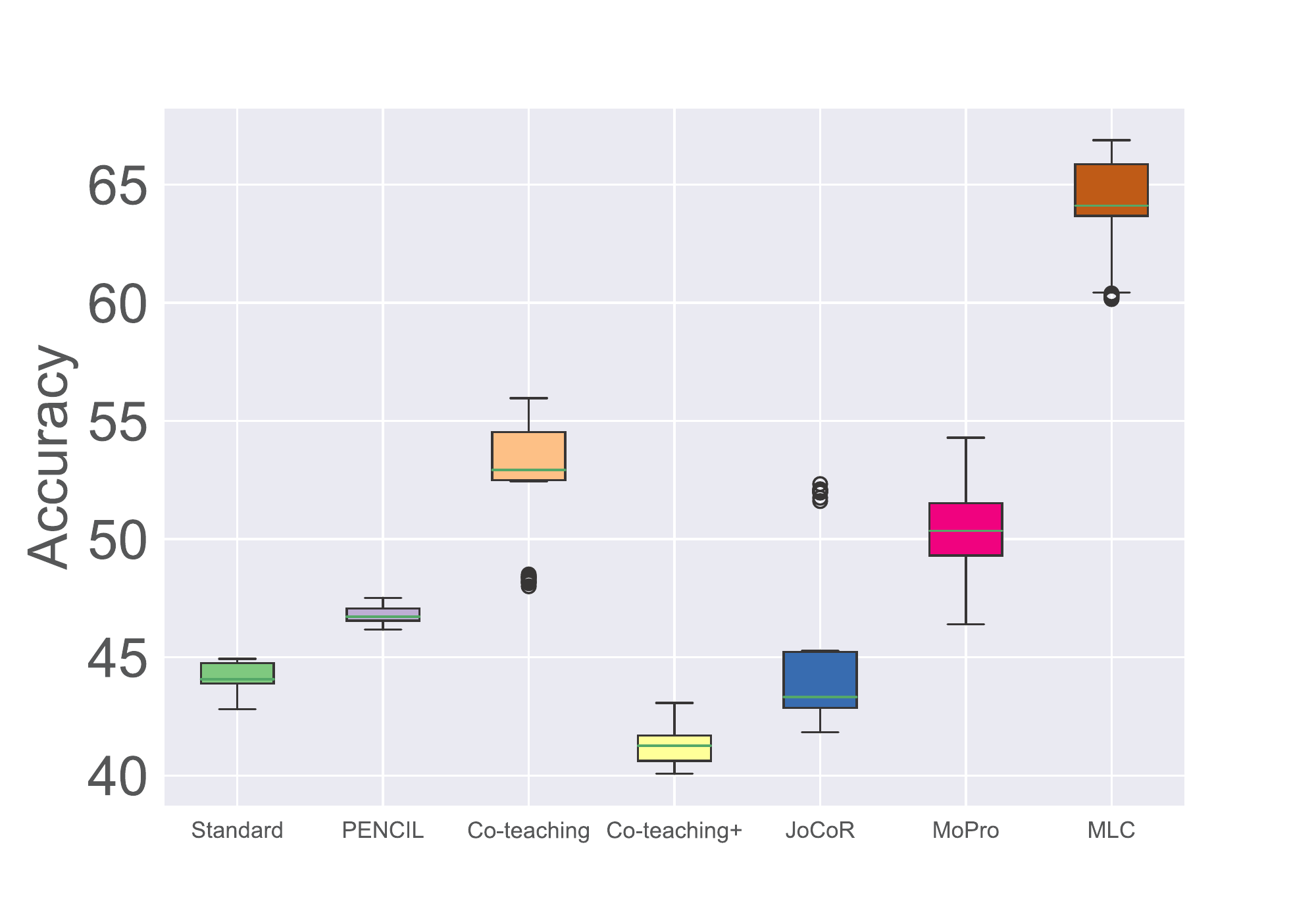} \label{fig:cifar10_box_b}
        }
        \subfigure[symmetric 0.2]{
        \includegraphics[width=4cm]{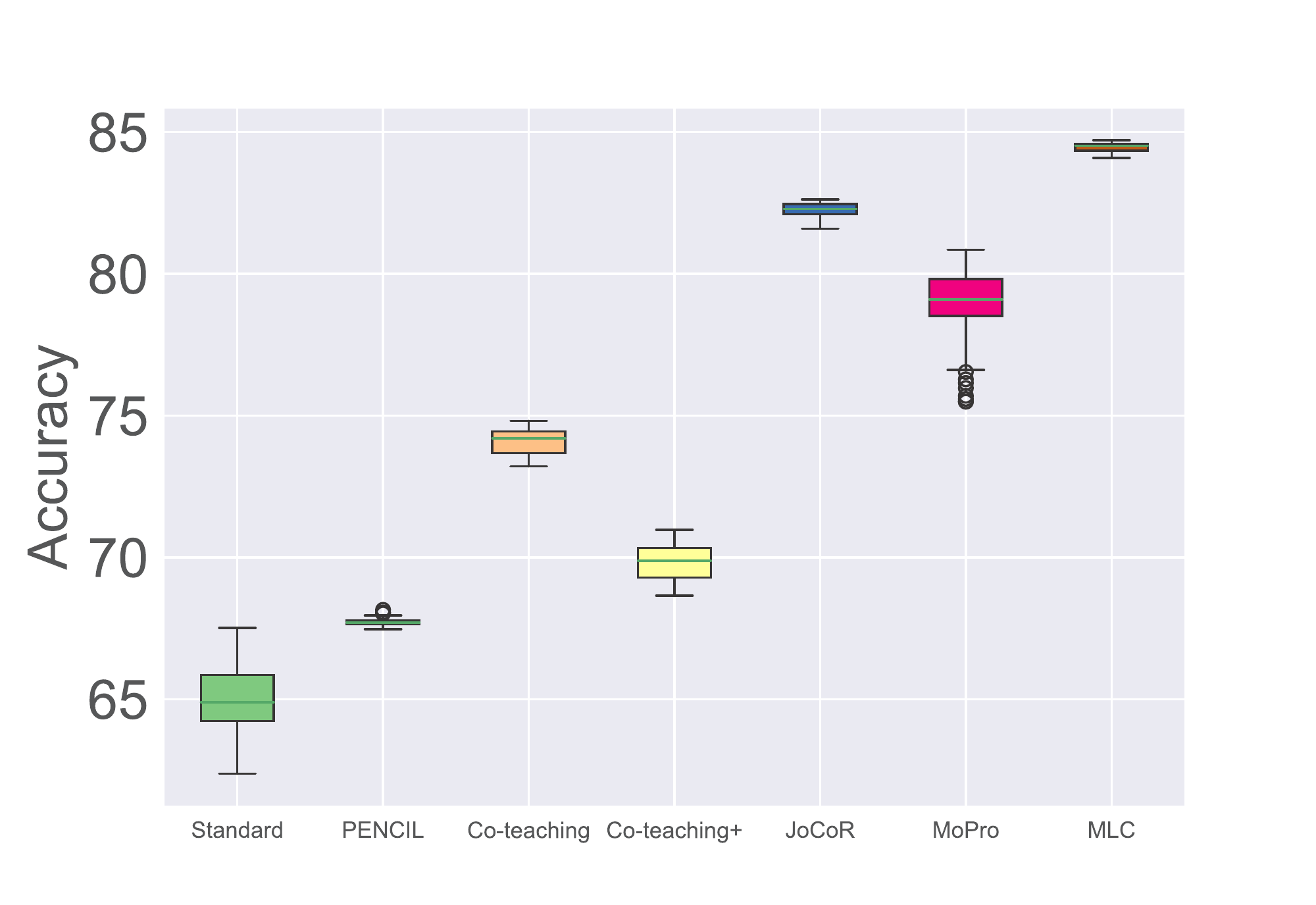} \label{fig:cifar10_box_c}
        }
        \subfigure[symmetric 0.4]{
        \includegraphics[width=4cm]{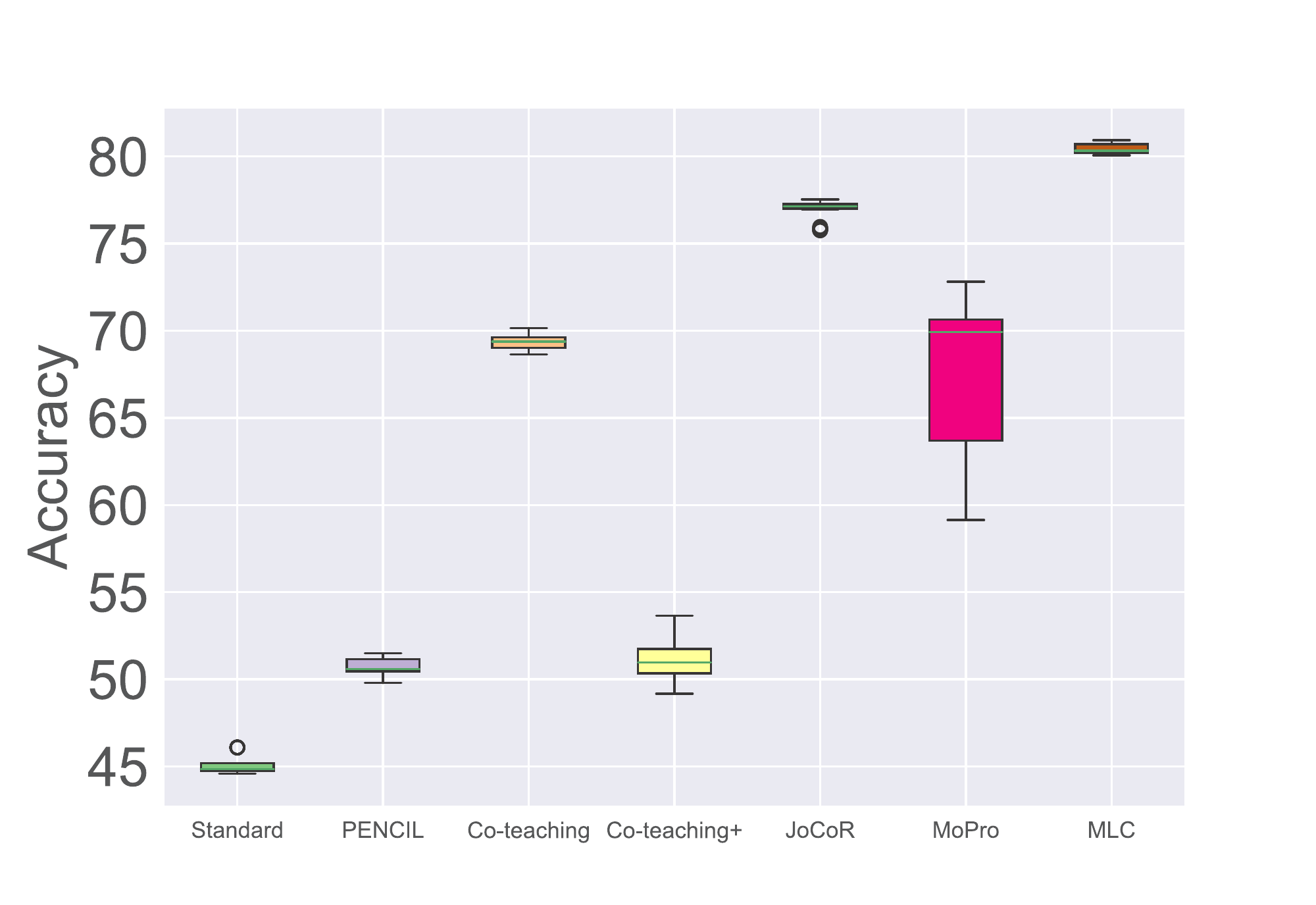} \label{fig:cifar10_box_d}
        }
        \subfigure[symmetric 0.8]{
        \includegraphics[width=4cm]{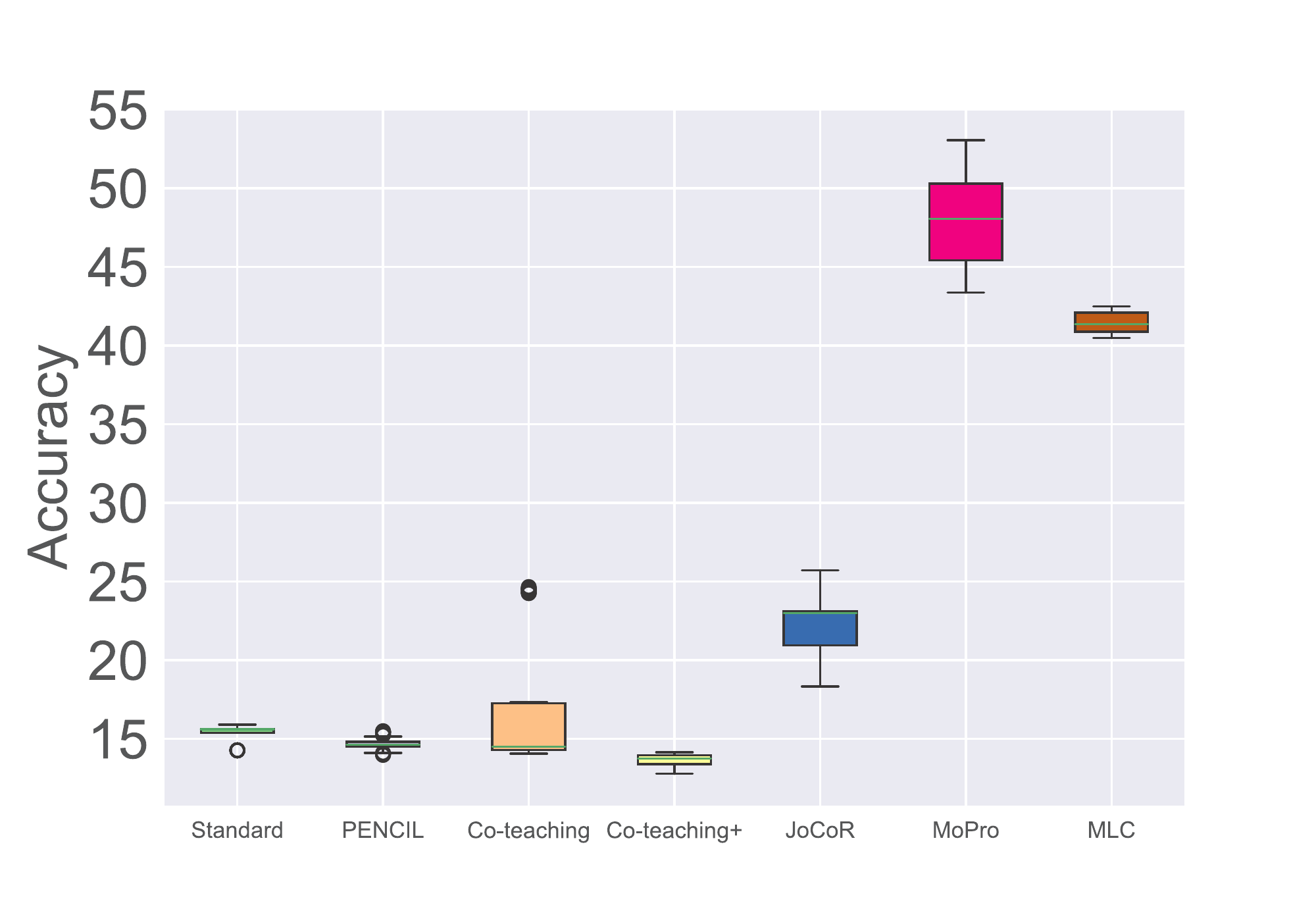} \label{fig:cifar10_box_e}
        }
        \hspace{4mm}
        \subfigure{
        \includegraphics[width=3cm]{figures/legend2.pdf}
        }
        \caption{Accuracy box-whisker plot of CIFAR-10 based on last ten epochs of five trials.}
        \label{fig:cifar10_box}
\end{figure}

Fig. \ref{fig:mnist_box} and Fig. \ref{fig:cifar10_box} show the accuracy (last) among five trials training results with various noise ratios and noise types of MNIST and CIFAR-10. MLC outperforms comparison methods by higher accuracy and more stable result. Surprisingly, the accuracy of Co-teaching is usually higher than Co-teaching+. In Fig. \ref{fig:mnist_box_a}, \ref{fig:mnist_box_c}, \ref{fig:mnist_box_d} and Fig. \ref{fig:cifar10_box_a}, \ref{fig:cifar10_box_c}, \ref{fig:cifar10_box_d}, the Bright yellow box (Co-teaching+) is lower and larger than the orange box (Co-teaching) and blue box (JoCoR), which means that Co-teaching and JoCoR are more accurate and stable than Co-teaching+ in most cases. \ADD{Specifically, MoPro has the higher accuracy in sn-0.8 of CIFAR-10 due to unsupervised feature learning but still suffers from instability.  In Fig.  \ref{fig:cifar10_box_d}, \ref{fig:cifar10_box_e}, the deep pink box (MoPro) is larger than the others. }\DEL{MLC has lower variance under a large noise ratio (Fig. \ref{fig:mnist_box_b}, \ref{fig:mnist_box_e} and Fig. \ref{fig:cifar10_box_b}, \ref{fig:cifar10_box_e}) since a certain disagreement is required for mutually training two networks, especially under a large noise ratio.}\ADD{On the contrary,  MLC has lower accuracy variance under a large noise ratio since a certain disagreement is required for mutually training two networks. For instance, in Fig. \ref{fig:mnist_box_b}, \ref{fig:mnist_box_e} and Fig. \ref{fig:cifar10_box_b}, \ref{fig:cifar10_box_e}, the sienna box (MLC) is smaller than the others. }

Interestingly, co-training-based methods are usually higher than others in the last accuracy, even if they are not guaranteed to obtain the best accuracy. For example, in sn-0.8 of MNIST, PENCIL and Standard won the second and third place of best accuracy. However, they were overtaken by the co-training-based methods over 40\% of the accuracy (last), which means they have fitted noisy labels. In fact, single network methods are capable of higher accuracy but easily memorize errors. In contrast, co-training-based methods are more robust. The gap between best and last accuracy is much lower than others, but there is still room to improve the best accuracy. By contrast. MLC obtain both best and last accuracy owing to better mutually training.

Another observation is that the label correction method PENCIL can be worse than Standard under pairflip noise. For instance, 0.45 pairflip noise in MNIST. That might is caused by inaccurate label correction, which is related to accurate class probability prediction. On the contrary, MLC performs better than others under pairflip noise with the help of mutual learning.

The result of Clothing1M (Table \ref{tab:acc_cloth}) is similar to MNIST and CIFAR-10. MLC gets the highest accuracy under real-world noise. Surprisedly, Co-teaching+ is worse than Standard in a large gap. The classification accuracy of MLC is slightly better than the two co-training methods and much higher than Co-teaching+.

\section{Conclusion}
This study set out to determine the necessity of “Disagreement” for training two networks to deal with noisy labels. The result shows that a certain degree of variance helps to judge label accuracy on noisy datasets better. Based on the observations, we designed a novel regularization term to ensure mutual collaboration between networks. Moreover, we establish a noise-tolerant framework named MLC with a mutual label correction mechanism and mutually regularization. MLC achieves the state-of-the-art result under different noise ratios/types of all datasets. In the future, it is expected that the proposed method can be adjusted to adapt to the needs of imbalanced data.

\section*{Acknowledgment}
This work was supported by the joint project of BRC-BC (Biomedical Translational) Engineering research center of BUCT-CJFH XK2020-07 and was supported in part by the National Natural Science Foundation of China under Grant No. 62101021.




\bibliographystyle{IEEEtran}
%
%
%

\bibliography{bibfile}

\end{document}